%% file: main.tex
\pdfoutput=1
\documentclass[acmtog,authorversion]{acmart}
\usepackage{booktabs} 
\usepackage{xcolor}
\usepackage{subcaption}
\usepackage{graphicx}

\usepackage{multirow}
\usepackage{amsmath}      

\newif\ifshowrevisions
\showrevisionsfalse
\usepackage[ruled]{algorithm2e} 

\SetAlFnt{\small}
\SetAlCapFnt{\small}
\SetAlCapNameFnt{\small}
\SetAlCapHSkip{0pt}

\usepackage{cleveref}

\crefname{section}{Sec.}{Secs.}
\Crefname{section}{Sec.}{Secs.}
\crefname{subsection}{Sec.}{Secs.}
\Crefname{subsection}{Sec.}{Secs.}

\crefname{equation}{Equation}{Equations}
\Crefname{equation}{Equation}{Equations}

\crefname{table}{Table}{Tables}
\Crefname{table}{Table}{Tables}

\crefname{figure}{Fig.}{Figs.}
\Crefname{figure}{Fig.}{Figs.}

\crefname{appendix}{Appendix}{Appendix}
\Crefname{appendix}{Appendix}{Appendix}
\acmJournal{TOG}
\copyrightyear{2026}
\acmYear{2026}
\acmConference[SIGGRAPH Conference Papers '26]{Special Interest Group on Computer Graphics and Interactive Techniques Conference Conference Papers}{July 19--23, 2026}{Los Angeles, CA, USA}
\acmBooktitle{Special Interest Group on Computer Graphics and Interactive Techniques Conference Conference Papers (SIGGRAPH Conference Papers '26), July 19--23, 2026, Los Angeles, CA, USA}
\acmDOI{10.1145/3799902.3811185}
\acmISBN{979-8-4007-2554-8/2026/07}

\begin{document}
\title{Learning Laplacian Eigenspace with Mass-Aware Neural Operators on Point Clouds}

\author{Zherui Yang}
\orcid{0009-0002-6928-6441}
\affiliation{%
  \institution{University of Science and Technology of China}
  \city{Hefei}
  \country{China}}
\email{zherui_yang@mail.ustc.edu.cn}

\author{Tao Du}
\orcid{0000-0001-7337-7667}
\affiliation{%
  \institution{Tsinghua University}
  \city{Beijing}
  \country{China}}
\affiliation{%
  \institution{Shanghai Qi Zhi Institute}
  \city{Beijing}
  \country{China}}
\email{taodu.eecs@gmail.com}

\author{Ligang Liu}
\orcid{0000-0003-4352-1431}
\authornote{Corresponding author.}
\affiliation{%
  \institution{University of Science and Technology of China}
  \city{Hefei}
  \country{China}}
\affiliation{%
  \institution{Laoshan Laboratory}
  \city{Qingdao}
  \country{China}}
\email{lgliu@ustc.edu.cn}

\begin{abstract}
The eigendecomposition of the Laplace--Beltrami Operator (LBO) is fundamental to geometric analysis, yet computing its low-frequency eigenmodes remains a significant bottleneck due to the high cost of iterative solvers on large-scale data.
To amortize this cost, we introduce the Neural Eigenspace Operator (NEO), a feed-forward framework designed to predict the spectrum directly from point clouds.
Crucially, NEO circumvents the ill-posed nature of standard eigenvector regression, which suffers from intrinsic sign flips and rotation ambiguities, by learning the stable, invariant low-frequency subspace instead.
Specifically, the network predicts a redundant set of basis functions whose span robustly covers the target eigenspace, allowing for the recovery of accurate eigenpairs via a lightweight Rayleigh--Ritz refinement.
To handle irregular sampling, we propose a mass-aware neural operator that incorporates per-point area weights into attention-based aggregation, improving robustness to non-uniform densities and enabling zero-shot generalization across resolutions.
Our approach achieves near-linear runtime scaling and substantial wall-clock speedups over iterative solvers at comparable accuracy, and exhibits strong zero-shot transfer to high-resolution point clouds.
The resulting eigenpairs support standard spectral geometry tasks, while the raw basis functions provide effective point-wise features for downstream learning.
Code: \url{https://github.com/Adversarr/NEO}.
\end{abstract}

%
%

\ccsdesc[500]{Computing methodologies~Shape analysis}
\ccsdesc[500]{Computing methodologies~Artificial intelligence}
\ccsdesc[300]{Computing methodologies~Neural networks}

%
%

\keywords{Spectral Geometry Processing, Eigenvalue Problem, Neural Operator}

\begin{teaserfigure}
  \includegraphics[width=\textwidth]{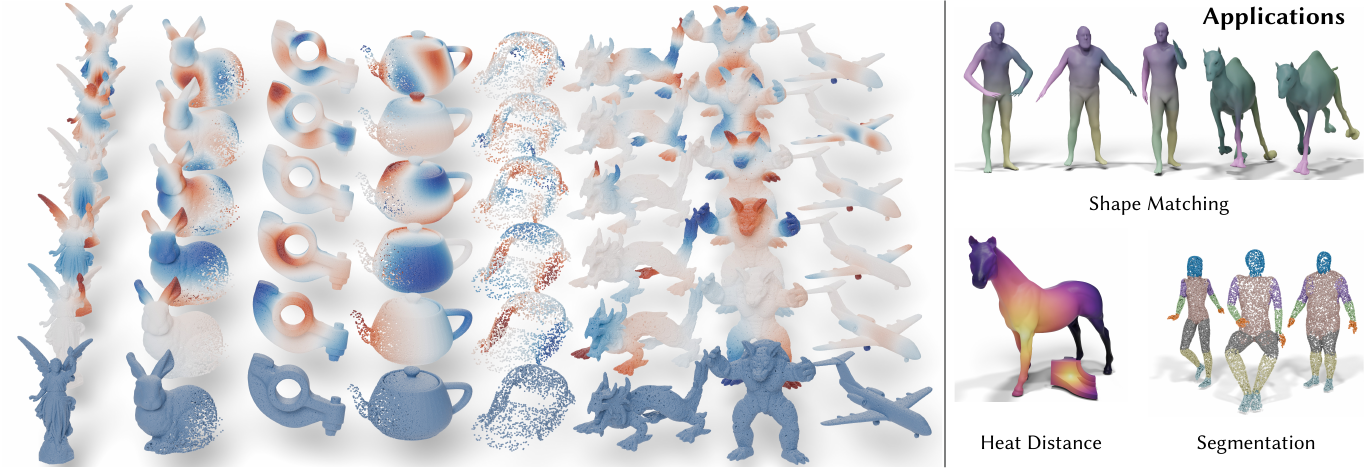}
  \caption{We present NEO, a neural framework that accelerates Laplace--Beltrami spectral analysis by predicting the low-frequency eigenspace directly from raw point clouds. \textbf{Left:} Example inputs with varying resolution, sampling density, and shape categories (bottom) and the corresponding predicted low-frequency eigenfunctions (top). \textbf{Right:} The resulting spectral representation can be used in downstream geometry processing tasks such as shape matching (functional maps), heat-method geodesics, and segmentation.}
  \Description{Teaser figure for NEO. The left side shows several input 3D point clouds with different resolutions, sampling densities, and shape categories, together with predicted low-frequency Laplace--Beltrami eigenfunctions visualized on the shapes. The right side illustrates downstream uses of the predicted spectrum, including functional-map shape matching, heat-based geodesic computation, and segmentation.}
  \label{fig:teaser}
\end{teaserfigure}

\maketitle
\input{p1-introduction}
\input{p2-related-works}
\input{p3-4-methods}
\input{p5-exp-app}

\section{Conclusion}
\label{sec:conclusion}
In this work, we present NEO, a feed-forward approach for low-frequency LBO eigensolving on 3D point clouds.
NEO predicts the low-frequency invariant eigenspace using a mass-aware neural operator, enabling fast spectral analysis and zero-shot transfer from low-resolution training to higher-resolution inputs in our experiments.
Across standard spectral pipelines, NEO provides consistent acceleration and exhibits near-linear runtime scaling with respect to the number of points when recovering low-frequency modes.

\begin{figure}[h]
    \centering
    \includegraphics[width=\linewidth]{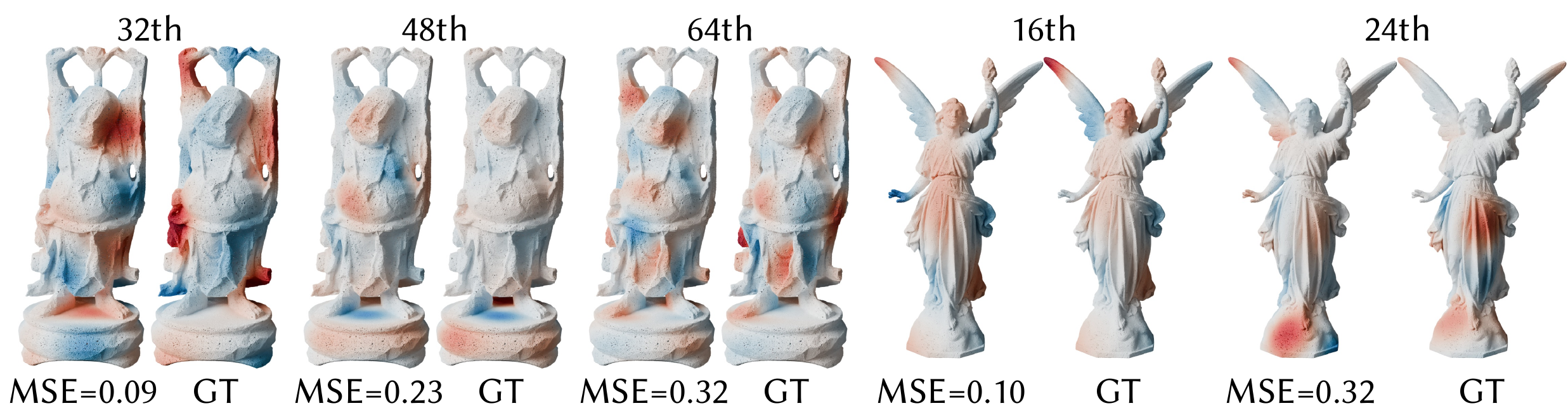}
    \caption{\textbf{Failure case.} We identify two main limitations: (1) Frequency degradation: Accuracy drops notably for higher modes, where the spectral gap narrows and oscillations become harder to approximate. (2) Unseen fine details: The model struggles with thin structures or complex topological details that differ significantly from the coarse geometry seen during training.}
    \Description{Failure-case figure showing two representative limitations of NEO. One example highlights degraded recovery for higher-frequency modes with denser oscillations, and another highlights errors on shapes with thin or fine-scale structures that differ from the coarse training geometry.}
    \label{fig:failure-case}
\end{figure}

\paragraph{Limitations and future work.}
As a learning-based method, NEO prioritizes inference speed over machine precision and is not intended to replace exact numerical solvers.
{Performance can degrade on higher modes and unseen thin structures (see \cref{fig:failure-case}).}
Promising future directions include using the predicted subspace to precondition iterative solvers and incorporating $SE(3)$-equivariance.

\begin{acks}
  This work was supported by Laoshan Laboratory (No. LSKJ202300305) and the National Natural Science Foundation of China (62025207).
  Tao Du acknowledges the support from Tsinghua University and the Shanghai Qi Zhi Institute Innovation Program.
  We also thank the anonymous reviewers for their valuable feedback and suggestions.
\end{acks}

\bibliographystyle{ACM-Reference-Format}
\bibliography{ref}

\section{Full Experiment Details}\label{app:full-experiments}

In this section, we provide the complete hyperparameter settings, training dynamics, and data generation details to ensure reproducibility.

\subsection{Model Architecture and Configurations}
\label{app:model_config}

\paragraph{Architecture Overview.}
The NEO backbone is instantiated using a Low-Rank Spatial Attention (LRSA) architecture. The pipeline begins with a coordinate lifting stage, followed by a stack of mass-aware operator blocks, and concludes with a linear projection to the output subspace functions.

\begin{itemize}
    \item \textbf{Input Encoding:} We map input coordinates $X \in \mathbb{R}^{N \times 3}$ to high-dimensional features using Sinusoidal Positional Encoding with frequencies logarithmically spaced in the range $[2^{-2}, 2^{6}]$. These are mapped to the model dimension $D$ via a linear layer with SiLU activation.
    \item \textbf{Latent-Bottleneck Blocks:} The core network consists of $L$ layers of LRSA blocks. Each block maintains a set of $r$ learnable latent tokens that act as a bottleneck for global information propagation. The block operates in three phases:
    (1) {Down-Projection} aggregates point features into the latent tokens via Cross-Attention;
    (2) {Latent Processing} mixes information among tokens via Self-Attention and Gated Feed-Forward Networks (FFN);
    (3) {Up-Projection} broadcasts the refined spectral information back to the point features via Cross-Attention.
    \item {Mass Injection:} To ensure robustness to non-uniform sampling, we explicitly modify the Down-Projection attention mechanism as described in the paper.
\end{itemize}

\paragraph{Model Variants.}
Table~\ref{tab:hyperparams} details the specific configurations used in our experiments.
NEO-Base (5.3M parameters) is our primary model.
NEO-Large is a deeper and wider variant used to demonstrate scaling laws.
NEO-Deep validates the sufficiency of our default depth.
NEO-Deep shows that additional depth can further improve subspace coverage, but NEO-Base offers the best accuracy–efficiency trade-off used throughout the paper.

\begin{table}[h]
\centering
\caption{\textbf{NEO Model Configurations.} All models target $k=96$ eigenpairs. Training time is measured on 4$\times$ A100 GPUs.}
\label{tab:hyperparams}
\small
\setlength{\tabcolsep}{4pt}
\begin{tabular}{l|ccc}
\toprule
{Parameter} & \textbf{NEO-Base} & \textbf{NEO-Large} & \textbf{NEO-Deep} \\
\midrule
\multicolumn{4}{l}{\textit{Architecture Specs}} \\
\quad Output Subspace ($m$)      & 192 & 256 & 192 \\
\quad Model Width ($D$)          & 192 & 256 & 192 \\
\quad Depth ($L$)                & 4   & 6   & 8   \\
\quad Attn. Heads                & 4   & 8   & 4   \\
\quad Attn. Head Dim             & 48  & 32  & 48  \\
\quad Params            & {5.3 M} & {16.9 M} & {10.6 M} \\
\midrule
\multicolumn{4}{l}{\textit{Training Statistics}}\\
\quad Training Time              & $\sim$8.6 hours & $\sim$1.5 days & $\sim$17.2 hours \\ 
\quad $\overline{\mathcal{E}}_{\mathrm{span}}$ (Train) & 3.07e-3 & 3.00e-3 & 1.97e-3\\
\quad $\overline{\mathcal{E}}_{\mathrm{span}}$ (Val) & 3.74e-3 & 3.59e-3 & 2.40e-3\\
\bottomrule
\end{tabular}
\end{table}

\subsection{Training and Data Pipeline}

\paragraph{Data Generation.}
Our training data is derived from the ShapeNetCore dataset. We generate the dataset by sampling each mesh twice independently with $N=2048$ points using surface-area-weighted sampling. Ground truth spectra are computed offline using ARPACK. 
This preprocessing step involves solving spectra for a large collection of shapes, consuming approximately 256 CPU-hours distributed across 64 cores.
We employ a strict 90/10 split for training and validation, ensuring no geometry leakage.

\paragraph{Optimization Setup.}
We optimize the model using the Muon~\cite{liu2025muonscalablellmtraining, jordan2024muon}
optimizer with momentum betas $(0.9, 0.99)$, a weight decay of $0.04$, $400$ epochs, and a maximum learning rate of $8\times 10^{-4}$.
To stabilize training dynamics, we apply gradient clipping with a threshold of $10.0$.
For learning rate scheduling, we adopt a {Warmup-Stable-Decay (WSD)} strategy:
(i) {Warmup:} Linearly increase LR for the first 10\% of steps;
(ii) {Stable:} Maintain constant LR for 70\% of steps;
(iii) {Decay:} Exponentially decay to a final LR of $1\times 10^{-6}$ over the final 20\%.
We use half-precision (float16) for the model forward pass. Although bfloat16 may be another common choice, we found that due to the low accuracy of bfloat16, the final training result is worse than float16.
In our experiments, AdamW~\cite{loshchilov2019decoupledweightdecayregularization} or OneCycle scheduler does not work better than Muon and WSD.

\paragraph{Normalization}
Regarding data preprocessing, we normalize all input point clouds to fit within the unit cube $[-1, 1]^3$ to ensure numerical stability across varying object scales. 
During training, we apply online data augmentation by performing random global rotations (sampled uniformly from $SO(3)$) on the input points. This strategy prevents overfitting to canonical poses and encourages the neural operator to learn rotation-robust geometric features.
 
\paragraph{Hardware and Computation.}
Models are trained on a single node equipped with 4$\times$ NVIDIA A100 (40GB) GPUs.
We utilize a per-GPU batch size of 200, resulting in a global effective batch size of 800. 
Under these settings, NEO-Base converges in approximately 8.6 hours. 
Scaling up the model increases costs linearly with depth: NEO-Deep ($L=8$) requires roughly 17 hours, while the wider NEO-Large ($L=6$) requires approximately 36 hours.

\section{Full Experiment Results}

\subsection{NEO as an Eigen Solver}
To strictly isolate the impact of architectural design from model capacity, we evaluate the PointNet++\cite{qi2017pointnet++} and Point Transformer \cite{zhao2021point} baselines under a rigorous iso-parameter setting. Specifically, we scale both baselines to match the parameter budget of NEO-Base ($\approx 5.3 \mathrm{M}$). Furthermore, to ensure a fair comparison, all variants are trained under a unified protocol, employing identical supervision (loss functions and regularization), optimization schedules, and batch configurations. This guarantees that any observed performance gaps are attributable solely to the architectural inductive biases.

As detailed in Table~\ref{tab:backbone}, while all models achieve reasonable convergence on the training resolution (2k), their behaviors diverge sharply under distribution shifts.
PointNet++ maintains moderate robustness but suffers from high latency due to its hierarchical sampling and grouping operations.
Crucially, Point Transformer exhibits catastrophic degradation when scaling to 64k points ($\mathcal{E}_{\mathrm{sub}}$ surges to $4.92\mathrm{e}{-}1$), indicating that standard local-attention mechanisms struggle to generalize zero-shot to density regimes unseen during training.
In contrast, NEO demonstrates superior stability across all settings, primarily due to its mass-aware global attention mechanism which effectively approximates resolution-invariant integral operators.
Furthermore, our architecture is significantly more efficient, delivering a $4.8\times$ to $8.9\times$ speedup over the baselines.

\begin{figure*}[t]
    \centering
    \includegraphics[width=1\linewidth]{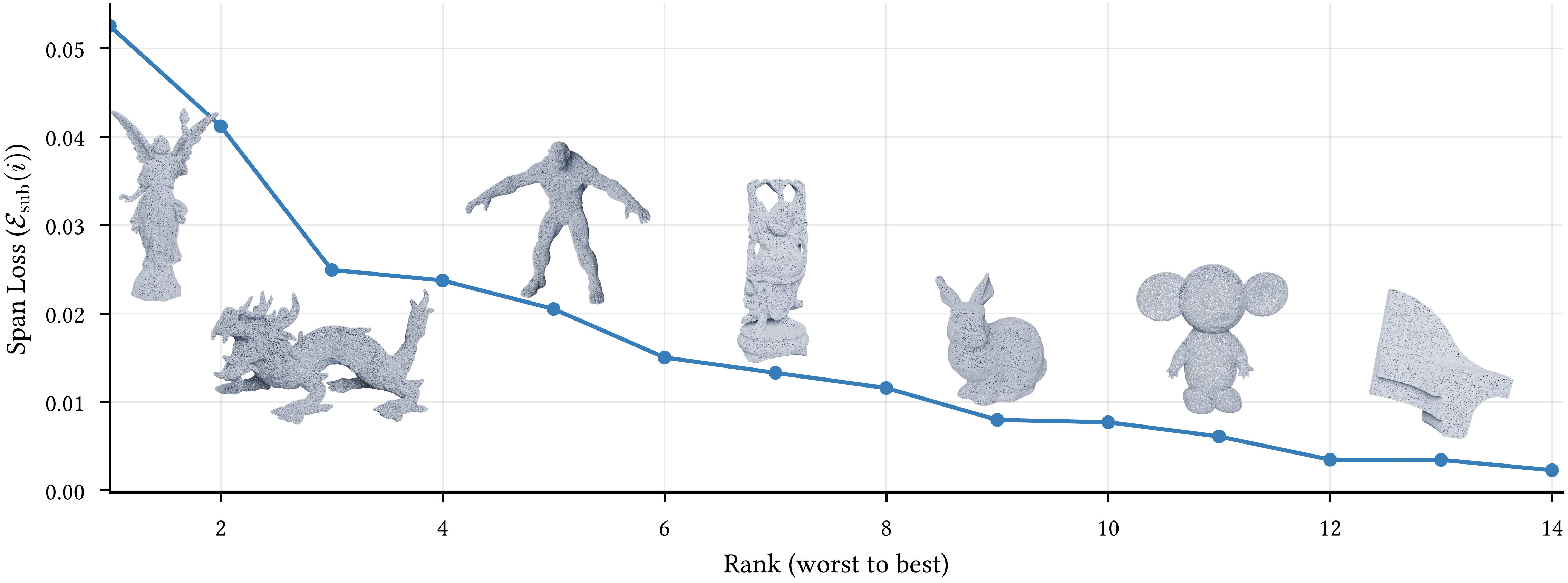}
    \caption{\textbf{Error distribution across OOD shapes ranked by Span Loss ($\mathcal{E}_{\mathrm{sub}}$).}
    We sort the OOD test shapes ($N=128\mathrm{k}$) from highest error (left) to lowest (right). 
    The results highlight a clear dependency on geometric complexity: performance degrades primarily on models with intricate high-frequency surface details or non-trivial topology, while remaining robust for smooth manifolds.}
    \label{fig:failure_case}
\end{figure*}

\begin{table}[h]
\centering
\caption{\textbf{Backbone Comparison details.} 
Performance is evaluated using the Mean Span Loss $\mathcal{E}_{\mathrm{sub}}$.
Baselines are parameter-matched to NEO-Base.
``Test(2k)'' evaluates the in-distribution test split at the training resolution; ``64k'' and ``Non-Uniform'' evaluate the OOD set.
$T_{\text{sample}}$ measures end-to-end backbone forward time (FP16) on a single RTX 3090.}
\label{tab:backbone}
\begin{tabular}{l|cccc}
\toprule
   \multirow{2}{*}{Backbone} &\multicolumn{3}{c}{${\mathcal{E}}_{\mathrm{sub}}$} & $T_{\text{sample}}$\\
    & Test(2k) & 64k & Non-Uniform & (ms) \\
\midrule
   PointNet++ & 6.97e-3 & 1.00e-2 & 1.21e-2 & 763\\
   Point Transformer  & 4.81e-3 & 4.92e-1 & 4.18e-1 & 1418 \\
   \textbf{NEO (Ours)} & \textbf{3.47e-3} & \textbf{3.27e-3} & \textbf{2.63e-3} & \textbf{159} \\ 
\bottomrule
\end{tabular}
\end{table}

\paragraph{Comparison with FastSpectrum.}
{We further compare NEO with FastSpectrum \cite{nasikun2018fastspectrum}, a mesh-based approximate eigensolver. 
Using the authors' implementation with matched subspace dimensions ($m=192$) and evaluation settings on our mesh datasets, we observed that while FastSpectrum offers a $14\times$ speedup over ARPACK (0.21s vs. 2.93s), NEO achieves a much more significant $73\times$ speedup (0.04s). 
As detailed in Table~\ref{tab:fastspectrum}, NEO also yields a lower mean Span Loss ($8.60\mathrm{e}{-}3$ compared to FastSpectrum's $1.08\mathrm{e}{-}2$) and lower Eigenvector MSE ($1.97\mathrm{e}{-}1$ vs. $2.28\mathrm{e}{-}1$), although with slightly higher variance ($1.17\mathrm{e}{-}1$ vs. $1.08\mathrm{e}{-}1$). 
This confirms that despite not assuming mesh connectivity, NEO predicts a high-quality spectral subspace and offers competitive or superior performance to specialized mesh-based approximation techniques.}

\begin{table}[h]
\centering
\caption{{\textbf{Comparison with FastSpectrum.} Evaluation on mesh datasets targeting $k=96$ eigenpairs with subspace dimension $m=192$. NEO yields lower span loss and eigenvector MSE but marginally higher variance.}}
\label{tab:fastspectrum}
\begin{tabular}{l|ccc}
\toprule
   Method & Span Loss & Evec. MSE & Evec. MSE Var. \\
\midrule
   FastSpectrum & $1.08\mathrm{e}{-}2$ & $2.28\mathrm{e}{-}1$ & $\mathbf{1.08\mathrm{e}{-}}1$ \\
   \textbf{NEO (Ours)} & $\mathbf{8.60\mathrm{e}{-}3}$ & $\mathbf{1.97\mathrm{e}{-}1}$ & $1.17\mathrm{e}{-}1$ \\
\bottomrule
\end{tabular}
\end{table}

\paragraph{Baselines Setting and Scalability Evaluation.}
{To comprehensively evaluate the computational efficiency and scalability of traditional eigensolvers across varying problem sizes, we benchmarked multiple industry-standard libraries: ARPACK, SPECTRA, and SLEPc (CUDA accelerated). We tested these solvers on matrices corresponding to point cloud discretizations ranging from $N = 8,192$ to over $1.5$ million points ($N = 1,572,864$). For SLEPc, we evaluated both the Shift-Invert and LOBPCG strategies. The experiments demonstrated that while traditional solvers like ARPACK and SPECTRA scale reasonably well at lower resolutions, their computational cost grows significantly for large-scale geometries. Notably, the SLEPc LOBPCG solver failed to converge or encountered memory/numerical issues under this specific configuration. Table~\ref{tab:solver_timings} summarizes the wall-clock times for each method.}

\begin{table}[ht]
\centering
\caption{
    {\textbf{Eigensolver Timing Comparison (seconds).} The evaluated time for ARPACK, SPECTRA, and SLEPc (Shift-Invert) across different problem sizes $N$. SLEPc with LOBPCG failed to evaluate under these settings.}
}
\label{tab:solver_timings}
\begin{tabular}{l|cccc}
\toprule
\textbf{Solver} & $N=\text{8k}$ & $N=\text{64k}$ & $N=\text{512k}$ & $N=\text{1.5M}$ \\
\midrule
ARPACK & 0.25 & 4.58 & 45.90 & 156.80 \\
Spectra & 0.34 & 5.65 & 61.01 & 208.66 \\
SLEPc & 0.48 & 5.21 & 63.71 & 249.08 \\
\textbf{NEO (FP32)} & 0.02 & 0.23 & 0.96 & 2.91\\
\textbf{NEO (FP16)} & 0.01 & 0.13 & 0.50 & 1.59\\
\bottomrule
\end{tabular}
\end{table}

\paragraph{Detailed Statistical Analysis.}
In Table~\ref{tab:metric_appendix_median_iqr}, we explicitly report the median and interquartile range (IQR) alongside the mean errors to analyze the error distribution.
Across all metrics, the median error is consistently lower than the mean (e.g., median $\overline{\mathcal{E}}_{\mathrm{sub}}$ is $15\%$ lower than the mean).
This discrepancy indicates a right-skewed error distribution, suggesting that NEO solves the vast majority of geometries with high precision, while the mean is slightly impacted by a small tail of outliers---likely shapes with severe non-manifold artifacts or disconnected components that challenge the definition of the ground-truth Laplacian itself.
The tight IQR confirms the reliability of our method across the diverse shape categories in ShapeNetCore.

\begin{table}[htbp]
\centering
\caption{\textbf{Accuracy Metrics on ShapeNet Test Split ($k=96$).} We report the mean, median, and interquartile range (IQR) across the test set. Span Loss is consistently low ($10^{-3}$), indicating robust coverage of the target low-frequency eigenspace.}
\label{tab:metric_appendix_median_iqr}
\begin{tabular}{c|ccc}
\toprule
   Metric  & Mean & Median & IQR \\
\midrule
   $\overline{\mathcal{E}}_{\mathrm{sub}}$ & 3.47e-03 & 2.84e-03 & 2.39e-03\\
   $\overline{\mathcal{E}}_{\mathrm{vec}}$ & 3.02e-02 & 2.67e-02 & 2.66e-02\\
   $\overline{\mathcal{E}}_{\mathrm{val}}$ & 3.61e-02 & 3.20e-02 & 2.25e-02\\
\bottomrule
\end{tabular}
\end{table}

\begin{figure}[h]
    \centering
    \includegraphics[width=1\linewidth]{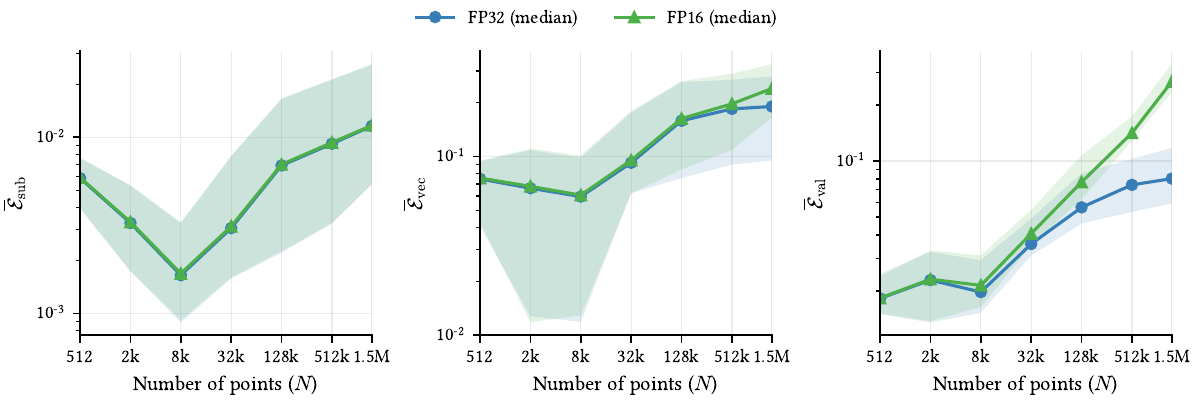}
    \caption{\textbf{Zero-shot resolution scalability and numerical precision.} We evaluate NEO, trained exclusively on 2k-point clouds, across a wide range of test resolutions ($N=512$ to $1.5\mathrm{M}$). Solid lines denote the median error; shaded regions correspond to the interquartile range (25--75\%).
\textbf{Left/Middle:} The subspace span loss ($\mathcal{E}_{\mathrm{sub}}$) and eigenvector MSE ($\mathcal{E}_{\mathrm{vec}}$) remain stable across resolutions, with the lowest error surprisingly occurring at $8\mathrm{k}$ points rather than the training resolution.
\textbf{Right:} While eigenvalues ($\mathcal{E}_{\mathrm{val}}$) are more sensitive to the precision gap at extremely high resolutions ($>128\mathrm{k}$), both eigenvector quality and subspace coverage are largely unaffected by the use of half-precision (FP16) inference.}
    \label{fig:ood_acc_var_n}
\end{figure}

\paragraph{Zero-shot Scalability and Precision Analysis.}
To validate the resolution scalability of our operator, we evaluate a single NEO model, which was trained strictly on $N_{\text{train}}\approx 2\mathrm{k}$, across OOD test point clouds ranging from $N=512$ to $1.5\times 10^6$.
As shown in Figure~\ref{fig:ood_acc_var_n}, the method exhibits strong zero-shot generalization.
Interestingly, the sweet spot for performance is observed at $N=8\mathrm{k}$, where both span loss and eigenvector MSE reach their minimum, outperforming even the native training resolution ($2\mathrm{k}$).
This suggests that the learned operator captures the underlying continuous spectral geometry better when the discretization granularity is slightly finer than training, likely because the discrete mass approximation becomes more accurate.
Performance degrades gracefully as $N$ increases beyond $10^5$, primarily due to the vast domain shift from the training distribution.
Regarding numerical precision, using FP16 for the neural backbone has a negligible impact on the predicted subspace ($\mathcal{E}_{\mathrm{sub}}$) and eigenvectors ($\mathcal{E}_{\mathrm{vec}}$).
However, the eigenvalue error ($\mathcal{E}_{\mathrm{val}}$) is more sensitive to numerical precision at very high resolutions ($N > 128\mathrm{k}$).
This behavior indicates that limited precision in the backbone output $F$ introduces high-frequency noise that is orthogonal to the target low-frequency subspace (keeping $\mathcal{E}_{\mathrm{vec}}$ low) but slightly perturbs the energy estimates in the projected Rayleigh quotient (inflating $\mathcal{E}_{\mathrm{val}}$), particularly as the conditioning of the Laplacian degrades with increasing $N$.

\paragraph{Failure Case Analysis}

Figure~\ref{fig:failure_case} visualizes the error distribution across the OOD test set at high resolution ($N=128\text{k}$), revealing a correlation between spectral accuracy and geometric complexity.
The most challenging cases correspond to models rich in intricate surface details (e.g., \emph{Dragon}, \emph{Lucy}), where dense micro-structures significantly perturb local area measures. Although our mass-aware mechanism mitigates global bias, capturing the aggregate spectral influence of these features remains difficult when they induce high-frequency localized modes.
Conversely, NEO achieves near-perfect reconstruction on piecewise smooth surfaces and simple topology (e.g., \emph{Bunny}, \emph{Fandisk}), confirming that the learned operator aligns well with the global low-frequency geometry of typical manifolds.

\paragraph{Accelerated Poisson Solver via Deflation} We provide the algorithm implemented in Section 6.1. (Accelerated geodesic distances).
The ICPCG baseline and ICPCG with deflation using NEO's predicted eigenpairs as presented in Alg.~\ref{alg:icpcg} and  Alg.~\ref{alg:deflated_icpcg}.

\begin{algorithm}[h]
\SetAlgoNoLine
\caption{Incomplete Cholesky PCG (ICPCG)}
\label{alg:icpcg}
\KwIn{System matrix $\mathbf{A}$, RHS $\mathbf{b}$, max iterations $K_{max}$, tolerance $\epsilon$}
\KwOut{Solution $\mathbf{x}$}

\tcp{Precompute Incomplete Cholesky Factorization}
$\mathbf{L} \leftarrow \text{IncompleteCholesky}(\mathbf{A})$\;
Define preconditioner solve $\mathcal{M}^{-1}(\mathbf{v}) \equiv (\mathbf{L}\mathbf{L}^T)^{-1}\mathbf{v}$\;
\BlankLine
$\mathbf{x}_0 \leftarrow \mathbf{0}$\;
$\mathbf{r}_0 \leftarrow \mathbf{b} - \mathbf{A}\mathbf{x}_0$\;
$\mathbf{z}_0 \leftarrow \mathcal{M}^{-1}(\mathbf{r}_0)$ \tcp*{Apply ICHOL preconditioner}
$\mathbf{p}_0 \leftarrow \mathbf{z}_0$\;
\BlankLine

\For{$k = 0$ \KwTo $K_{max}$}{
    $\alpha_k \leftarrow (\mathbf{r}_k^T \mathbf{z}_k) / (\mathbf{p}_k^T \mathbf{A} \mathbf{p}_k)$\;
    $\mathbf{x}_{k+1} \leftarrow \mathbf{x}_k + \alpha_k \mathbf{p}_k$\;
    $\mathbf{r}_{k+1} \leftarrow \mathbf{r}_k - \alpha_k \mathbf{A} \mathbf{p}_k$\;
    
    \If{$\|\mathbf{r}_{k+1}\| < \epsilon$}{
        \textbf{break}\;
    }
    
    $\mathbf{z}_{k+1} \leftarrow \mathcal{M}^{-1}(\mathbf{r}_{k+1})$ \tcp*{Apply ICHOL preconditioner}
    
    $\beta_k \leftarrow (\mathbf{r}_{k+1}^T \mathbf{z}_{k+1}) / (\mathbf{r}_k^T \mathbf{z}_k)$\;
    $\mathbf{p}_{k+1} \leftarrow \mathbf{z}_{k+1} + \beta_k \mathbf{p}_k$\;
}
\Return $\mathbf{x}_{k+1}$\;
\end{algorithm}

\begin{algorithm}[h]
\SetAlgoNoLine
\caption{Spectral Deflated ICPCG (Additive)}
\label{alg:deflated_icpcg}
\KwIn{System matrix $\mathbf{A}$, RHS $\mathbf{b}$, Basis vectors $\mathbf{Y} \in \mathbb{R}^{N \times k}$} from NEO
\KwOut{Solution $\mathbf{x}$}

\tcp{1. Setup Deflation Space}
$\mathbf{E} \leftarrow \mathbf{Y}^T \mathbf{A} \mathbf{Y}$ \tcp*{Compute coarse matrix $k \times k$}
$\mathbf{E}^{-1} \leftarrow \text{Inverse}(\mathbf{E})$ \tcp*{Dense inverse for small $k$}
Define coarse solve $\mathcal{C}(\mathbf{v}) \equiv \mathbf{Y} (\mathbf{E}^{-1} (\mathbf{Y}^T \mathbf{v}))$\;

\tcp{2. Setup Base Preconditioner (High-Freq)}
$\mathbf{L} \leftarrow \text{IncompleteCholesky}(\mathbf{A})$\;
Define base solve $\mathcal{M}_{base}^{-1}(\mathbf{v}) \equiv (\mathbf{L}\mathbf{L}^T)^{-1}\mathbf{v}$\;
\BlankLine

\tcp{3. Warm Start using coarse solution}
$\mathbf{x}_0 \leftarrow \mathcal{C}(\mathbf{b})$\;
$\mathbf{r}_0 \leftarrow \mathbf{b} - \mathbf{A}\mathbf{x}_0$\;

\tcp{Two-level additive apply: $\mathbf{z} = \mathbf{z}_{base} + \mathbf{z}_{coarse}$}
$\mathbf{z}_0 \leftarrow \mathcal{M}_{base}^{-1}(\mathbf{r}_0) + \mathcal{C}(\mathbf{r}_0)$\; 
$\mathbf{p}_0 \leftarrow \mathbf{z}_0$\;
\BlankLine

\For{$k = 0$ \KwTo $K_{max}$}{
    $\alpha_k \leftarrow (\mathbf{r}_k^T \mathbf{z}_k) / (\mathbf{p}_k^T \mathbf{A} \mathbf{p}_k)$\;
    $\mathbf{x}_{k+1} \leftarrow \mathbf{x}_k + \alpha_k \mathbf{p}_k$\;
    $\mathbf{r}_{k+1} \leftarrow \mathbf{r}_k - \alpha_k \mathbf{A} \mathbf{p}_k$\;
    
    \If{$\|\mathbf{r}_{k+1}\| < \epsilon$}{
        \textbf{break}\;
    }
    
    \tcp{Additive Preconditioner Step}
    $\mathbf{z}_{high} \leftarrow \mathcal{M}_{base}^{-1}(\mathbf{r}_{k+1})$\;
    $\mathbf{z}_{low} \leftarrow \mathcal{C}(\mathbf{r}_{k+1})$\;
    $\mathbf{z}_{k+1} \leftarrow \mathbf{z}_{high} + \mathbf{z}_{low}$\;
    
    $\beta_k \leftarrow (\mathbf{r}_{k+1}^T \mathbf{z}_{k+1}) / (\mathbf{r}_k^T \mathbf{z}_k)$\;
    $\mathbf{p}_{k+1} \leftarrow \mathbf{z}_{k+1} + \beta_k \mathbf{p}_k$\;
}
\Return $\mathbf{x}_{k+1}$\;
\end{algorithm}

\subsection{NEO as a Point-wise Embedding Provider}
We investigate the utility of the raw predicted functions $F \in \mathbb{R}^{N \times m}$ as intrinsic point embeddings for downstream tasks.
In this ``Laplacian-free'' inference mode, NEO serves as a pre-trained geometric feature extractor.
To rigorously evaluate the quality of these embeddings, we compare them against a standard positional encoding baseline (NeRF-PE) combined with either a PointNet or Point Transformer head.

\paragraph{Data Processing and Discretization.}
Since the original datasets (SHREC'11 and Human Body) provide triangular meshes, we convert them to point clouds to simulate a raw scanning setup.
For classification, we uniformly sample $N=8192$ points from the mesh surface.
For segmentation, we perform training and inference on the sampled point cloud; to evaluate against the ground-truth labels defined on the mesh vertices, we propagate the predicted labels from the point cloud to the original vertices via nearest-neighbor interpolation.

\paragraph{Implementation Details.}
To ensure that performance differences arise from the input features rather than model capacity, we employ standardized lightweight heads for all baselines.
Table~\ref{tab:app_hyperparams} details the specific hyperparameters for the PointNet++ and Point Transformer architectures used in these experiments.
All variants share a consistent feature dimension ($d=128$) and activation scheme.
For the NEO-based method, we freeze the backbone and only train the task-specific head (a simple PointNet MLP) on top of the predicted features $F$.

\paragraph{Positional Encoding.}
To establish a competitive coordinate-based baseline, we employ the Positional Encoding (PE) mechanism popularized by NeRF~\cite{mildenhall2020nerf}. 
Since standard MLPs exhibit a spectral bias towards low-frequency signals~\cite{rahaman2019spectral}, raw coordinates $\mathbf{x} \in \mathbb{R}^3$ are insufficient for capturing fine-grained geometric details. NeRF-PE lifts the input into a higher-dimensional Fourier feature space via 
\begin{equation*}
\gamma(\mathbf{x}) = [\sin(2^0\pi\mathbf{x}), \cos(2^0\pi\mathbf{x}), \dots, \sin(2^{L-1}\pi\mathbf{x}), \cos(2^{L-1}\pi\mathbf{x})],
\end{equation*}
allowing the network to regress high-frequency variations effectively.
In our experiments, we set $L=10$, resulting in a 63-dimensional input vector (including original coordinates).

\begin{table}[htbp]
\centering
\caption{\textbf{Hyperparameters for Downstream Task Baselines.} 
We detail the architectures used for the few-shot classification and segmentation heads.
Both models share the same channel width ($d=128$) and general depth ($L=4$ or 3 stages) where applicable. "SA" denotes Set Abstraction and "FP" denotes Feature Propagation modules.}
\label{tab:app_hyperparams}
\begin{tabular}{ll}
\toprule
\textbf{Hyperparameter} & \textbf{Value} \\
\midrule
\textit{PointNet++} \\
\quad Base Width ($d_{\text{model}}$) & 128 \\
\quad Activation & SiLU \\
\quad SA Downsampling Ratios & $[0.2, 0.25]$ \\
\quad SA Ball Query Radii & $[0.2, 0.4]$ \\
\quad SA MLP Channels & $[128, 256, 1024]$ \\
\quad FP Up-sampling $k$-NN & $[1, 3, 3]$ \\
\quad Lifting Layers & 1 (Linear-Act-Linear) \\
\midrule
\textit{Point Transformer} \\
\quad Base Width ($d_{\text{model}}$) & 128 \\
\quad Depth (Layers) & 4 \\
\quad Local Neighborhood ($k$) & 16 \\
\quad Lifting Blocks & 1 (Linear-Act-Linear) \\
\quad Normalization & LayerNorm \\
\quad Activation & SiLU \\
\bottomrule
\end{tabular}
\end{table}

\paragraph{Few-shot Classification.}
Table~\ref{tab:cls_fewshot} reports the accuracy on the SHREC'11 few-shot classification benchmark.
Our method ($F$ + PointNet) significantly outperforms coordinate-based baselines in the low-data regime.
Notably, in the 1-shot setting, NEO achieves an accuracy of $58.3\%$, surpassing the Point Transformer baseline by over $16\%$.
This suggests that the spectral subspace encapsulates a robust geometric prior that generalizes better than raw coordinates when supervision is scarce.

\begin{table}[h]
\centering
\caption{\textbf{Few-shot classification on SHREC'11.} 
We report the classification accuracy (\%) for 1, 3, and 10-shot settings. 
The comparison includes our method (NEO) and coordinate-based baselines (NeRF-PE) equipped with different heads. 
``Mass Free'' denotes the ablation where integration weights are fixed to $w_i=1$.}
\begin{tabular}{l|cccc}
\toprule
Method / Shot & 1 & 3 & 10 \\
\midrule
NEO ($F$) + PointNet & \textbf{58.3} & \textbf{84.2} & \textbf{100.0} \\
NEO ($F$) + PointNet (Mass Free) & 53.3 & 80.8 & 99.1 \\
NeRF-PE + PointNet & 41.7 & 69.2 & 95.8 \\
NeRF-PE + Point Transformer & 47.5 & 82.5 & 96.7 \\
\bottomrule
\end{tabular}
\label{tab:cls_fewshot}
\end{table}

\paragraph{Dense Segmentation.}
Table~\ref{tab:seg_miou} evaluates dense prediction performance on the Human Body segmentation task.
We observe that using NEO features leads to faster convergence and higher final accuracy.
At 100 epochs, our method reaches near-optimal performance (0.80 mIoU), whereas coordinate-based methods require significantly more training steps to align the spatial information.
This confirms that $F$ provides a globally consistent coordinate system that simplifies the optimization landscape for dense labeling tasks.

\begin{table}[h]
\centering
\caption{\textbf{Segmentation on Human Body.} 
We report the mean IoU (mIoU) evaluated at 100 and 300 training epochs. 
All methods utilize the same frozen backbone settings and a lightweight PointNet head. 
The comparison highlights the convergence speed and final performance differences between spectral and coordinate-based features.}
\begin{tabular}{l|cc}
\toprule
Method / Epochs & 100 & 300 \\
\midrule
NEO ($F$) + PointNet & \textbf{0.80} & \textbf{0.82} \\
NEO ($F$) + PointNet (Mass Free) & 0.78 & 0.80 \\
NeRF-PE + PointNet & 0.77 & 0.80 \\
NeRF-PE + Point Transformer & 0.78 & 0.80 \\
\bottomrule
\end{tabular}
\label{tab:seg_miou}
\end{table}

\paragraph{Ablation on Mass Awareness in Downstream Tasks.}
We further analyze the contribution of the mass integration weights by including a ``Mass Free'' variant ($w_i=1$) in both Table~\ref{tab:cls_fewshot} and Table~\ref{tab:seg_miou}. 
This setting simulates a simplified deployment scenario where local area estimation is skipped, treating the integration effectively as a uniform sum.
Results indicate that while the mass-aware formulation consistently yields the best performance, the method exhibits {graceful degradation} when mass is omitted.
In the classification task (Table~\ref{tab:cls_fewshot}), the gap is most pronounced in the 1-shot setting (58.3\% vs. 53.3\%), suggesting that accurate geometric integration is crucial for extracting robust priors when supervision is scarce. 
However, it is notable that even the Mass Free variant significantly outperforms the NeRF-PE baseline (53.3\% vs. 47.5\%), confirming that the learned spectral subspace captures intrinsic manifold structure that raw coordinates cannot, regardless of the integration scheme.

\end{document}

%% file: p1-introduction.tex

\section{INTRODUCTION}


Eigenmodes of the Laplace--Beltrami operator (LBO) provide an intrinsic spectral representation of 3D shapes \cite{levy2006laplace}.
Much like the Fourier basis on Euclidean domains, the \emph{low-frequency} eigenfunctions capture global, smooth geometric structures, underpinning diverse graphics pipelines~\cite{vallet2008spectral, ovsjanikov2012functional}, from spectral mesh processing \cite{levy2010spectral, zhang2007spectral} and physical simulation \cite{pentland1989good} to geometric deep learning \cite{bronstein2017geometric, sharp2022diffusionnet}.

In practice, obtaining this basis reduces to solving a sparse generalized eigenvalue problem (GEVP) arising from an LBO discretization.
While discretization is typically local and efficient, extracting the first $k$ low-frequency modes remains a primary computational bottleneck.
Standard eigensolvers based on Krylov subspaces (e.g., implicitly restarted Lanczos) compute these modes through iterative procedures.
As a result, the computation is inherently \emph{per-instance}: changes in resolution, re-sampling, or deformation often require recomputing the decomposition.
This cost is prohibitive at scale, pushing spectral analysis to offline preprocessing in many pipelines.


\begin{figure}
    \centering
    \includegraphics[width=1\linewidth]{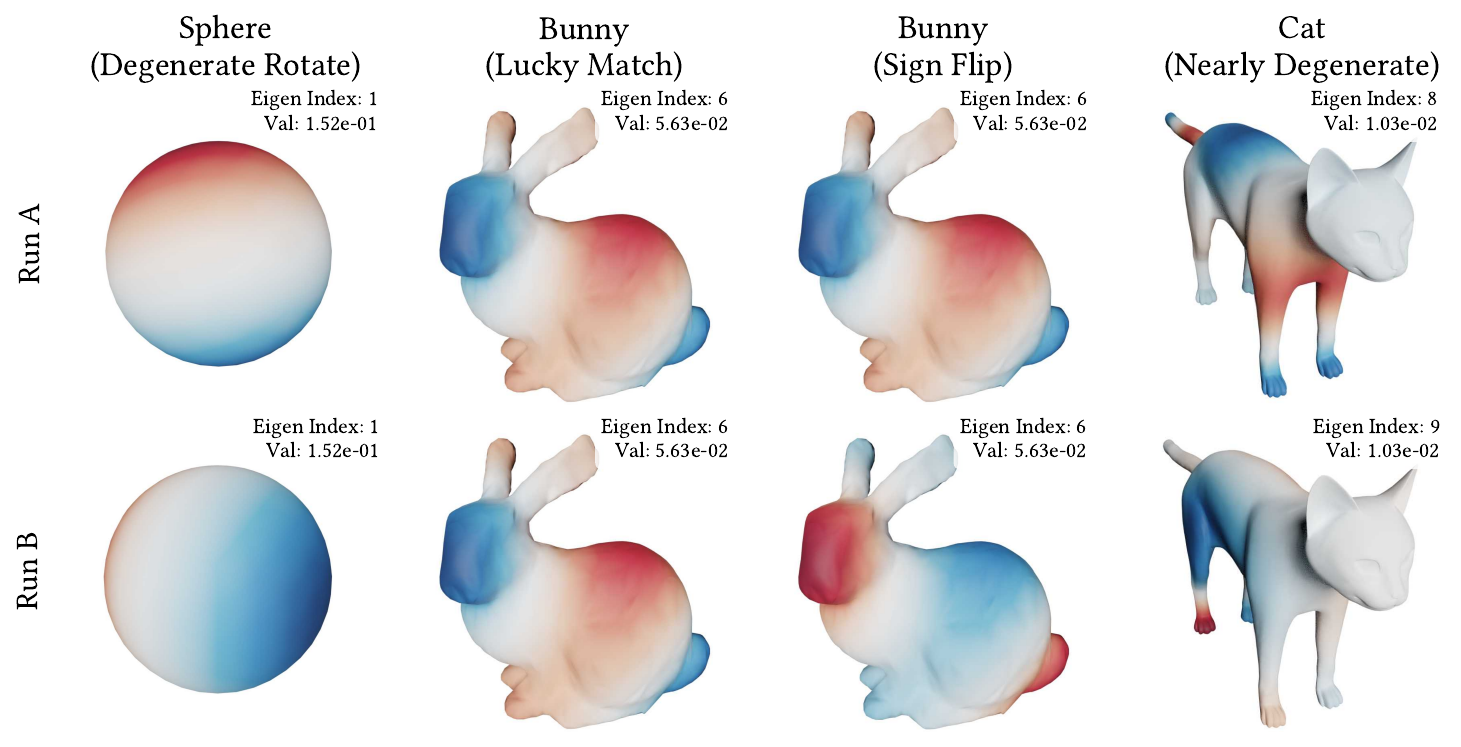}
    \caption{\textbf{Ambiguities in LBO eigenspace computation.}
Eigenfunctions are not uniquely defined: each mode is determined only up to a global sign, and repeated (or nearly repeated) eigenvalues admit arbitrary orthogonal bases within the same eigenspace.
Consequently, equally valid eigensolvers (or the same solver under small numerical perturbations) may return different eigenvector bases for the same shape (A vs.\ B), including (left) rotations within a degenerate eigenspace, (middle) sign flips, and (right) mode mixing or re-ordering near spectral clusters.
These ambiguities render direct eigenvector regression ill-posed, as there is no unique basis to align with.
}
    \Description{Comparison figure illustrating ambiguities in Laplace--Beltrami eigenvectors for the same shape. The examples show valid solutions that differ by a rotation within a repeated eigenspace, a sign flip of an eigenfunction, or mode mixing and reordering near a spectral cluster.}
    \label{fig:intro-ambiguity}
\end{figure}


Motivated by the success of feed-forward pipelines such as RenderFormer~\cite{zeng2025renderformer} and VGGT~\cite{wang2025vggt}, we generalize this paradigm to spectral analysis by learning to solve the GEVP directly on 3D point clouds.
Our goal is to predict a high-quality approximation of the low-frequency LBO eigenspace in a single feed-forward pass.
However, directly regressing the discretized eigenvectors is ill-posed due to inherent ambiguities~\cite{davis1970rotation, golub2013matrix, chang2024shape}: each eigenvector is defined only up to a sign flip, and repeated eigenvalues admit arbitrary rotations within the associated eigenspace (see Fig.~\ref{fig:intro-ambiguity}).
Consequently, regressing individual vectors forces the network to memorize the arbitrary basis choices in the training data, leading to unstable training and poor generalization.
To address this, we leverage the fact that while individual eigenfunctions are ambiguous, the low-frequency invariant subspace they span is unique and well-defined.
We therefore shift the learning objective from regressing individual eigenfunctions to predicting a set of functions whose span robustly captures the target eigenspace.

To this end, we introduce the \emph{Neural Eigenspace Operator} (NEO), a feed-forward framework that predicts the low-frequency invariant subspace from raw point clouds.
Rather than outputting exactly $k$ modes, NEO predicts a modestly redundant set of $m$ functions ($m>k$) whose span robustly captures the target eigenspace, providing slack to accommodate repeated or nearby eigenvalues.
Crucially, to handle the irregularity of point cloud sampling, we employ a \emph{mass-aware} attention mechanism in the neural operator. 
This module explicitly incorporates point masses into feature aggregation, ensuring the neural operator approximates the continuous integral operator with respect to the underlying measure, rather than being biased by sampling density.
We train NEO with a rotation-invariant loss that compares spans rather than individual eigenvectors, directly resolving the aforementioned eigenvector ambiguities.
When explicit eigenpairs are required, we apply a lightweight {Rayleigh--Ritz refinement}. 
This step projects the discretized
operators onto the predicted low-dimensional basis, reducing the original large-scale sparse GEVP to a small dense eigenproblem that is efficient to solve.


Our experiments show that NEO achieves near-linear inference scaling, accelerating eigensolving on point clouds ranging from thousands to over a million points.
At the same time, it preserves spectral accuracy by predicting subspaces with low span error and supporting reliable Rayleigh--Ritz eigenpair recovery.
Thanks to its mass-aware neural operator, NEO can be trained exclusively on low-resolution data yet exhibits strong zero-shot transfer to dense geometry, while remaining robust under non-uniform sampling and varied discretizations.
The recovered eigenpairs can be used in standard spectral geometry pipelines and the raw predicted functions also provide effective point-wise features for downstream learning tasks.
We summarize our contributions as follows:
\begin{itemize}
\item \textbf{Neural Eigensolving Framework.} 
We propose a learning-based framework that reformulates low-frequency LBO eigensolving as invariant subspace prediction followed by Rayleigh-Ritz refinement.
This approach enables rapid linear-time inference after one-time training.
\item \textbf{Geometric Learning Formulation.}
We integrate a mass-aware mechanism into the neural operator, so that it respects the surface measure and improves robustness to non-uniform sampling.
Together with a rotation-invariant span loss, these formulations render the ill-posed eigenvector regression task feasible and stable.
\end{itemize}

%% file: p2-related-works.tex
\section{RELATED WORK}

\subsection{LBO Eigensolvers}
Classical LBO eigendecomposition relies on sparse eigensolvers for large-scale GEVPs.
Krylov-subspace algorithms, such as the implicitly restarted Lanczos method in ARPACK~\cite{lehoucq1998arpack}, and block methods such as LOBPCG~\cite{knyazev2001toward}, are widely adopted due to their robustness and accuracy.
In practice, their performance can be substantially improved with preconditioning and shift-invert schemes, especially when high-precision low-frequency modes are required.
{To improve efficiency, geometric approximation methods~\cite{nasikun2018fastspectrum} construct explicit subspaces, but typically rely on mesh connectivity. Nevertheless, both classical sparse eigensolvers and these approximation schemes remain per-instance: they solve each GEVP independently and cannot exploit statistical patterns across a collection of shapes to amortize inference.}

Recent advances in physics-informed machine learning offer optimization-based alternatives.
Physics-informed neural networks (PINNs)~\cite{raissi2019physics} and variational methods such as Deep Ritz~\cite{yu2018deepritz, ben2023deep} optimize continuous objectives to recover eigenfunctions.
Recent work further extends these ideas to parameterized families of geometries and studies how spectra vary across a shape space~\cite{chang2024shape}.
While these methods are appealing when differentiability with respect to geometry is needed, they generally still require solving an optimization problem for each new instance.
In contrast, we target one-shot prediction of the low-frequency invariant subspace, which can be optionally refined by a small projected solve.

\subsection{Neural Operators}
Neural operators learn mappings between function spaces~\cite{kovachki2023neural}, offering a resolution-agnostic paradigm well suited for geometric spectral problems where shapes and eigenfunctions are represented as point-sampled fields.
Seminal works like DeepONet~\cite{lu2021learning} and Fourier Neural Operators (FNO)~\cite{li2020fourier} have demonstrated their efficacy on regular grids.
For irregular geometries, graph-based approaches~\cite{li2020multipole, pfaff2020learning} provide local message passing, yet scaling them to capture long-range correlations over dense point sets often requires deep stacks or global mechanisms~\cite{alon2020bottleneck}.
To enable more efficient global interactions, methods such as Transolver~\cite{wu2024transolver} and LinearNO~\cite{hu2025lineartransolver} route information using attention through a small set of latent tokens, achieving near-linear scaling in the number of input points.
NEO adopts this scalable paradigm and adapts it by incorporating mass-aware aggregation.
{This design is reminiscent of surface attention~\cite{trappolini2023shape}, introduced for shape correspondence. In contrast, NEO injects mass information into the attention score in order to obtain a resolution-invariant integral approximation in the neural-operator setting.}
Concurrently, \citet{li2025den} solve parametric non-selfadjoint eigenvalue problems by learning invariant subspaces.
While their approach focuses on spectral analysis under varying physical parameters over fixed discretizations, our work emphasizes geometric variability across shapes and employs a mass-aware mechanism to maintain resolution-invariance on raw point clouds.

\subsection{Spectral Geometry \& Geometric Learning}
Spectral methods provide an intrinsic representation for geometry processing and shape analysis, rooted in manifold learning and spectral shape analysis~\cite{belkin2003laplacian, coifman2006diffusion, reuter2006laplace}.
Low-frequency LBO eigenfunctions underpin widely used descriptors and bases, including HKS/WKS~\cite{sun2009concise, aubry2011wave} and spectral bases for correspondence via functional maps and their refinements~\cite{ovsjanikov2012functional, melzi2019zoomout, ren2018continuous}.
In geometric learning, spectral constructions have likewise informed model design, from early spectral convolutions~\cite{Bruna2013SpectralNA, Yi2016SyncSpecCNNSS} to recent architectures that use low-frequency eigenvectors for feature propagation on surfaces~\cite{sharp2022diffusionnet, Smirnov2021HodgeNet}.
In most such pipelines, eigenpairs are assumed to be precomputed.
{Neural Laplacian Operator~\cite{pang2024neural} predicts a high-quality discrete Laplacian operator directly from point clouds, enabling downstream computations via standard eigensolvers. NeuralSound~\cite{jin2022neuralsound} similarly uses subspace learning, but is designed for fast modal sound synthesis on voxel discretizations. In contrast, NEO directly predicts the eigenspace on raw point clouds, providing fast access to spectral bases commonly used in geometry processing.}

%% file: p3-4-methods.tex
\section{NEO INFERENCE PIPELINE}
\label{sec:pipeline}


\label{sec:method_overview}
\begin{figure*}[t]
    \centering
    \includegraphics[width=\linewidth]{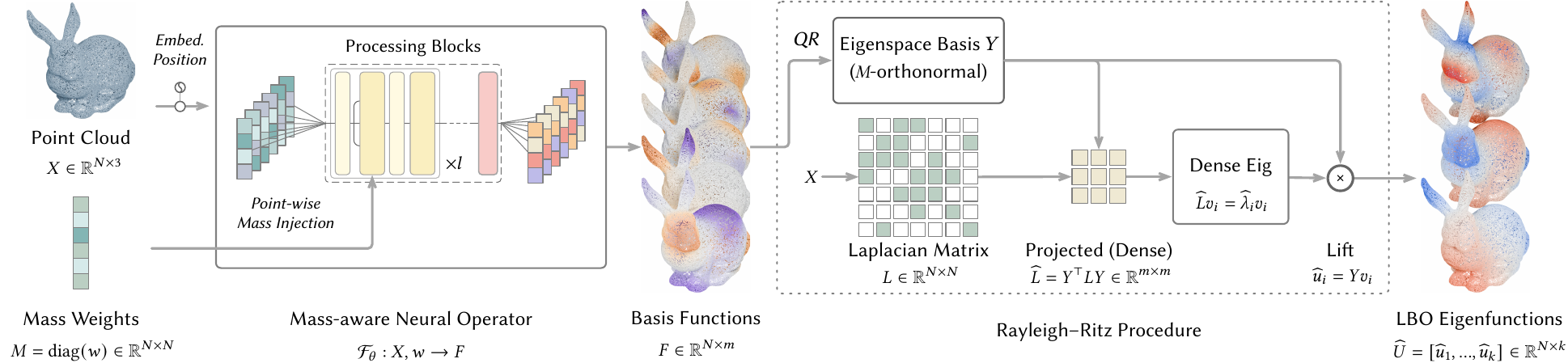}
    \caption{\textbf{Overview of the NEO Inference Pipeline.} Given a point cloud $X$ with mass weights $w$ (defining $M$), a mass-aware neural operator predicts redundant raw basis functions $F$ in one forward pass. 
    We $M$-orthonormalize $F$ to obtain a subspace basis $Y$, project the discrete Laplacian to a small dense matrix $\widehat{L}=Y^\top L Y$, solve a dense eigenproblem in the subspace, and lift the eigenvectors back as $\widehat{u}_i = Y v_i$.}
    \Description{Pipeline overview of NEO. Starting from a point cloud and per-point mass weights, the neural operator predicts a redundant set of functions. These functions are orthonormalized with respect to the mass matrix to form a subspace basis, the Laplacian is projected into this low-dimensional subspace, a small dense eigenproblem is solved, and the resulting eigenvectors are lifted back to the original points.}
    \label{fig:method-pipeline}
\end{figure*}



In this section, we present the inference pipeline of NEO.
We first formalize the discrete spectral problem on point clouds (\cref{subsec:setup}).
Then, we detail how NEO predicts a low-frequency invariant subspace to recover explicit eigenpairs via Rayleigh--Ritz refinement, which also yields intrinsic point embeddings as a byproduct (\cref{subsec:pipeline}).
The training objectives and network architecture are detailed in \cref{sec:training}.

\subsection{Problem Setup}\label{subsec:setup}

Let $\mathcal{M}$ be a (possibly bounded) Riemannian manifold and $\Delta$ its Laplace--Beltrami operator.
We consider the eigenvalue problem $-\Delta \phi_i = \lambda_i \phi_i$ with $0=\lambda_0 \le \lambda_1 \le \dots$, and assume Neumann boundary conditions for manifolds.
The eigenfunctions $\{\phi_i\}$ form an orthonormal basis of $L^2(\mathcal{M})$ under the inner product $\langle f,g\rangle_{\mathcal{M}} = \int_{\mathcal{M}} f(x)g(x)\,d\mu(x)$.
In this work, we focus on the invariant subspace $\mathcal{S}_k = \mathrm{span}(\{\phi_i\}_{i=0}^{k-1})$ spanned by the first $k$ eigenfunctions.

We operate on raw point clouds $X = \{x_i\}_{i=1}^N \subset \mathbb{R}^3$ sampled from $\mathcal{M}$ without assuming mesh connectivity.
To define the discrete spectral problem, we assume access to a sparse, symmetric positive semi-definite (PSD) Laplacian matrix $L\in\mathbb{R}^{N\times N}$ and a \emph{diagonal} mass matrix $M=\mathrm{diag}(w)\succ 0$ that discretize the continuous operators (e.g., the intrinsic Delaunay construction~\cite{sharp2020laplacian}).
Here, $w_i>0$ approximates the local area measure at $x_i$, inducing the discrete inner product $\langle u,v\rangle_M = u^\top M v$.
Functions on $\mathcal{M}$ are thus represented by vectors $u\in\mathbb{R}^N$ sampled at the input points, and the eigenproblem is then discretized into the GEVP:
\begin{equation}
    L u_i = \lambda_i M u_i,
    \qquad u_i^\top M u_j = \delta_{ij},
    \label{eq:gevp}
\end{equation}
Our goal is to replace this costly eigensolving for the first $k$ modes with a single feed-forward evaluation.


\subsection{The NEO Pipeline}
\label{subsec:pipeline}


{As shown in \cref{fig:method-pipeline},} to circumvent eigenvector ambiguities, NEO predicts the \emph{invariant subspace} $\mathcal{S}_k$ itself, which is mathematically unique.
At inference time, the mass-aware neural operator $\mathcal{F}_\theta$ maps the input point cloud to a set of $m$ basis functions:
\begin{equation}
    F=\mathcal{F}_\theta(X, w)\in\mathbb{R}^{N\times m},
\end{equation}
whose span is intended to cover $\mathcal{S}_k$.
We deliberately introduce redundancy $m > k$ as a relaxation: this allows the predicted subspace $\mathrm{span}(F)$ to robustly cover the target subspace even in the presence of ambiguities, without being forced to resolve them inside the network.
Since $m \ll N$, this representation remains compact even for high-resolution point clouds.
When explicit eigenpairs are needed, we refine within this subspace via Rayleigh--Ritz, as described next.

\paragraph{Eigenpair recovery via Rayleigh--Ritz refinement.}
For applications requiring explicit eigenpairs $(\lambda_i, u_i)$, we apply a standard Rayleigh--Ritz procedure for the GEVP within the predicted subspace.
First, we convert the raw fields $F$ into an $M$-orthonormal basis $Y\in\mathbb{R}^{N\times m}$ without changing the span.
Since $M$ is diagonal, we can efficiently compute a weighted QR factorization: let $Z=\sqrt{M}\,F$, compute its Euclidean QR factorization $Z=QR$, and set
\begin{equation}
Y = M^{-1/2}Q,\qquad \text{so that } Y^\top M Y = I_m .
\label{eq:m_qr}
\end{equation}
We then project the LBO onto the $m$-dimensional subspace:
\begin{equation}
    \widehat{L} = Y^\top L Y \in \mathbb{R}^{m\times m},
    \qquad
    \widehat{M} = Y^\top M Y = I_m .
    \label{eq:project_operators}
\end{equation}
The original high-dimensional GEVP is thus reduced to a small, dense $m \times m$ eigen-decomposition:
\begin{equation}
    \widehat{L} v_i = \widehat{\lambda}_i v_i,
    \qquad
    \widehat{u}_i = Y v_i .
    \label{eq:uplift}
\end{equation}
We take the $k$ smallest Ritz pairs $\{(\widehat{\lambda}_i,\widehat{u}_i)\}_{i=0}^{k-1}$ as the recovered low-frequency eigenmodes.
Overall, this replaces a large sparse GEVP with: (i) one network forward pass,
(ii) an $M$-orthonormalization ($\mathcal{O}(N m^2)$),
(iii) a sparse--dense projection ($\mathcal{O}(\mathrm{nnz}(L)\,m^2)$),
and (iv) a dense eigen-decomposition ($\mathcal{O}(m^3)$).
Assuming the discrete Laplacian sparsity is proportional to $N$ (e.g., constant valence), the total inference cost scales linearly with $N$.

\paragraph{Direct embedding.}
For downstream learning tasks (e.g., classification or segmentation), explicit eigenpairs are often unnecessary.
As a byproduct of predicting a low-frequency subspace, we can use the raw basis $F\in \mathbb{R}^{N\times m}$ directly as point-wise features.
This usage is \emph{Laplacian-free} at inference time: it requires only the point coordinates (and optionally mass weights), bypassing both Laplacian assembly and Rayleigh--Ritz refinement, and incurring no additional cost beyond a single forward pass.
For simplicity, in these downstream tasks we set $w_i=1$ when sampling is close to uniform.


\section{TRAINING AND ARCHITECTURE}
\label{sec:training}

In this section, we open the black box of NEO and describe how the neural operator $\mathcal{F}_\theta(X, w)$ in \cref{subsec:pipeline} is learned in practice.
We first define the subspace loss that enables supervised learning from ambiguous spectral data. 
We then detail the architecture of the neural operator, focusing on how the mass-aware design facilitates zero-shot generalization to arbitrary sampling densities.

\subsection{Distance Measure between Linear Spaces}
\label{subsec:loss}


We assume access to a collection of training tuples $\{(X, w, U_k)\}$, where $U_k$ represents the ground-truth $M$-orthonormal eigenbasis.
Our objective is to maximize the overlap between the predicted subspace $\mathrm{span}(F)$ and the ground-truth eigenspace $\mathrm{span}(U_k)$.


To this end, we minimize the residual energy of the ground-truth modes after projection onto the predicted subspace.
Let $Y \in \mathbb{R}^{N \times m}$ ($m \ge k$) be the $M$-orthonormal basis obtained from the network output $F$ via $M$-orthogonalization.
The $M$-orthogonal projector onto the predicted subspace is given by $P_Y = Y Y^\top M$.
For a specific ground-truth eigenvector $u_j$, the portion of energy not captured by the predicted subspace is:
\begin{equation}
    r_j = \|u_j - P_Y u_j\|_M^2 = 1 - \|Y^\top M u_j\|_2^2,
\end{equation}
where the simplification holds because $\|u_j\|_M = 1$.
Intuitively, minimizing $r_j$ forces the network to align its output span with the directions of maximum spectral energy, without constraining it to match any specific rotation or sign of the basis vectors.
Aggregating over all $k$ modes, we define the \textbf{span loss} as:
\begin{equation}
    \mathcal{L}_{\text{span}} = \frac{1}{k}\sum_{j=1}^k \left(1 - \|Y^\top M u_j\|_2^2\right) = 1 - \frac{1}{k}\|Y^\top M U_k\|_F^2 .
    \label{eq:span_loss}
\end{equation}
Since $\|Y^\top M u_j\|_2^2$ measures the energy of $u_j$ projected onto $\mathrm{span}(Y)$, the condition $\mathcal{L}_{\text{span}}=0$ implies that each mode $u_j$ is perfectly contained within the predicted span, i.e., $\mathrm{span}(U_k) \subseteq \mathrm{span}(Y)$.
Moreover, $\mathcal{L}_{\text{span}}$ is invariant under any orthogonal change of basis $U_k \mapsto U_k R$ for $R\in O(k)$, which includes sign flips and rotations within (near-)degenerate eigenspaces.

While the span loss ensures coverage, the redundancy $m > k$ can theoretically allow the raw fields $F$ to collapse into a rank-deficient state.
To encourage full-rank diversity in the raw predictions, we add a weak orthogonality regularizer:
\begin{equation}
    \mathcal{L}_{\text{ortho}} = \left\|F^\top M F - I_m\right\|_F^2 .
    \label{eq:ortho_loss}
\end{equation}
Overall, we optimize
$\mathcal{L}_{\text{total}} = \mathcal{L}_{\text{span}} + \alpha\,\mathcal{L}_{\text{ortho}},$
with {$\alpha$ kept fixed at $10^{-3}$} across all experiments.

\subsection{Mass-Aware Neural Operator}
\label{subsec:backbone}
\begin{figure}[t]
    \centering
    \includegraphics[width=1\linewidth]{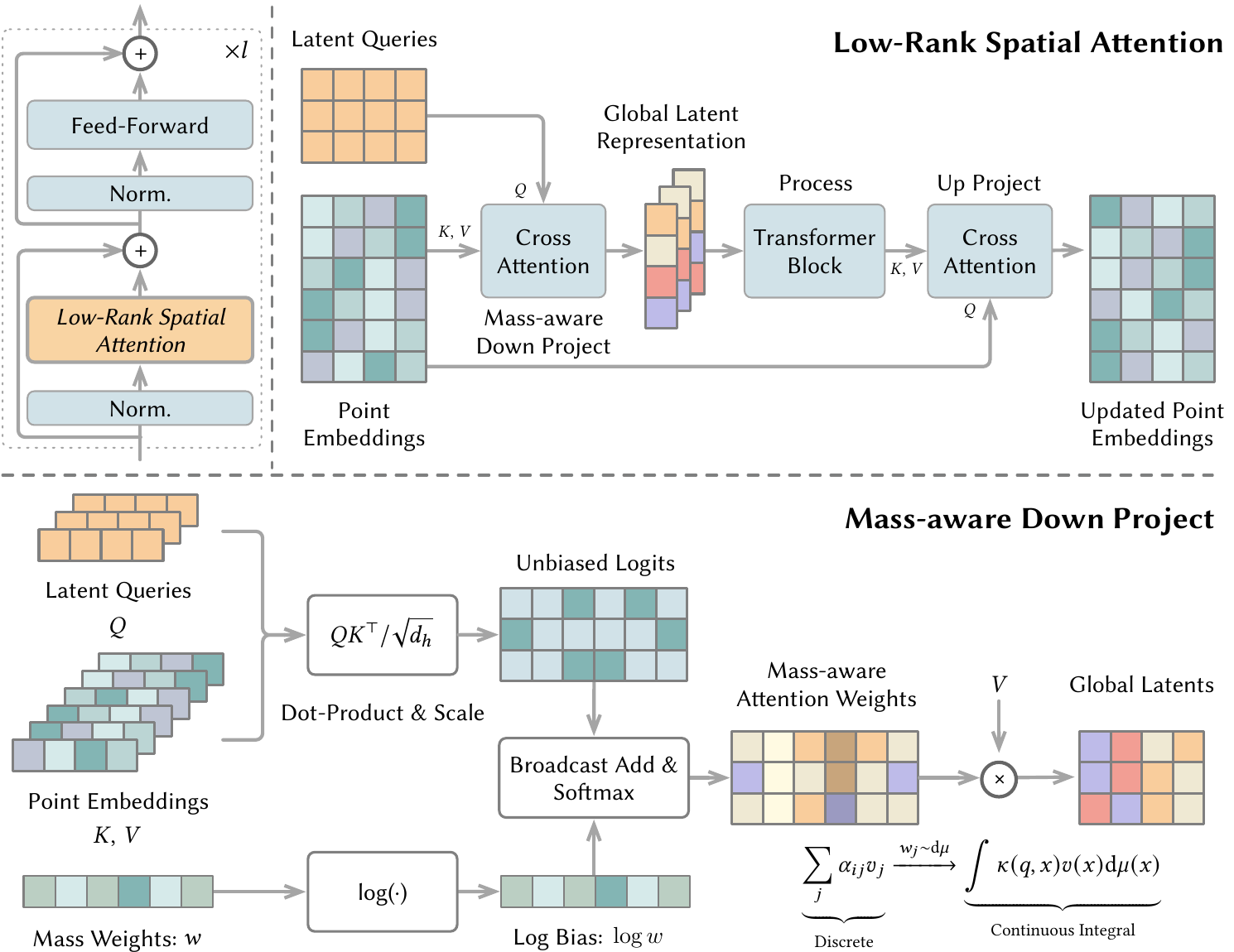}
\caption{\textbf{Neural Operator with Mass Injection.}
The architecture (top) adopts a latent-bottleneck design to efficiently process high-resolution geometry via global tokens.
To ensure resolution invariance, the down-projection (bottom) incorporates point masses via a logarithmic bias in the cross-attention, approximating a continuous measure-weighted integral.}
\Description{Architecture diagram of the neural operator backbone. The top part shows a latent-bottleneck design that maps point features to a smaller set of global latent tokens and back. The bottom part highlights the modified cross-attention used in down-projection, where point masses are injected as a bias so aggregation approximates a measure-weighted integral and remains stable under non-uniform sampling.}
\label{fig:backbone}
\end{figure}

To enable resolution-invariant inference, we instantiate the neural operator $\mathcal{F}_\theta$ with a latent-bottleneck transformer backbone, adapting the Low-Rank Spatial Attention (LRSA) architecture \cite{yang2026simpleeffectivelowrankspatial}.
Given an input point cloud $X\in\mathbb{R}^{N\times 3}$ and its associated mass weights $w\in\mathbb{R}^N_{>0}$, we first construct point-wise embeddings by lifting coordinates into a feature space:
\begin{equation}
h_i^{(0)}=\mathrm{MLP}_{\mathrm{lift}}([\mathrm{PE}(x_i)]),\qquad
H^{(0)}=[h_1^{(0)},\ldots,h_N^{(0)}]^\top\in\mathbb{R}^{N\times d},
\end{equation}
where $\mathrm{PE}(\cdot)$ denotes a sinusoidal positional encoding and $\mathrm{MLP}_{\mathrm{lift}}$ is applied independently to each point.
LRSA then updates the point embeddings through $l$ pre-norm transformer blocks, where the global interaction is implemented by a low-rank latent bottleneck (Fig.~\ref{fig:backbone} top): point embeddings are softly aggregated into a compact set of latent tokens via cross-attention, processed in the latent space, and broadcast back to all points via a second cross-attention.
A final point-wise linear projection maps the updated embeddings to $m$ output scalar fields, $F\in\mathbb{R}^{N\times m}$, which serve as raw basis functions for the subsequent operations.
However, standard cross-attention aggregation implicitly assumes uniform sampling; we explicitly address this via a \emph{mass injection} mechanism, as derived next.


\paragraph{Mass injection in cross-attention.}
In LRSA, the down-projection aggregates point-wise features into a compact set of latent vectors $\{z_i\}_{i=1}^r$ (with $r \ll N$) via cross-attention, forming a global latent representation.
Let $Q \in \mathbb{R}^{r \times d}$ be a set of learnable latent queries, and let $K, V \in \mathbb{R}^{N \times d}$ be the keys and values derived from the point embeddings $H$.
With a positive kernel $\kappa(q_i, k_j) \doteq \exp(\langle q_i, k_j\rangle / \sqrt{d_h})$, standard attention produces a normalized unweighted average:
\begin{equation}
\label{eq:attn_discrete}
z_i = \sum_{j=1}^N \underbrace{\frac{\kappa(q_i, k_j)}{\sum_{n=1}^N \kappa(q_i, k_n)}}_{\alpha _{ij}} v_j 
\;\approx\; 
\frac{\int_{\mathcal{M}} \kappa(q_i, k(x)) v(x) d\mu(x)}{\int_{\mathcal{M}} \kappa(q_i, k(x)) d\mu(x)}.
\end{equation}
The approximation on the right holds \emph{only when} the discretization is uniform, so that each sample represents a comparable area.

For geometric inputs with highly non-uniform sampling, Eq.~\eqref{eq:attn_discrete} can over-emphasize densely sampled regions and weaken generalization across resolutions.
To correct this, we inject the point masses $w$ directly into the attention logits.
By adding $\log w$ to the logits, we transform the aggregation into a consistent \emph{quadrature rule}:
\begin{equation}
\label{eq:attn_mass}
\begin{aligned}
\alpha'_{ij} &= \mathrm{softmax}\left(\frac{q_i k_j^\top}{\sqrt{d_h}} + \log w_j\right)_j
= \frac{\kappa(q_i, k_j) \cdot w_j}{\sum_{n=1}^N \kappa(q_i, k_n) \cdot w_n}, \\
z_i' &= \sum_{j=1}^N \alpha'_{ij} v_j 
\;\approx\; 
\frac{\int_{\mathcal{M}} \kappa(q_i, k(x)) v(x) d\mu(x)}{\int_{\mathcal{M}} \kappa(q_i, k(x)) d\mu(x)}.
\end{aligned}
\end{equation}
Since $\exp(s+\log w_j)=\exp(s)\,w_j$, the mass term emerges as a multiplicative quadrature weight outside the exponential kernel, making the aggregation consistent with area-weighted integration on non-uniform point clouds.
Importantly, this modification is invariant to the global scale of $w$ and exactly recovers the standard attention of Eq.~\eqref{eq:attn_discrete} when the mass weights are constant.
We apply this correction only in the down-projection, as it is the stage that aggregates point samples over $\mathcal{M}$; the up-projection attends from points to latent tokens, and latents are not samples of $\mathcal{M}$ and therefore do not admit geometric mass weights.

%% file: p5-exp-app.tex
\section{EXPERIMENTS}
\label{sec:experiments}
In this section, we evaluate NEO as a fast spectral solver.
Complete architecture details, training hyperparameters, and evaluation statistics are provided in Appendix A.

\begin{figure*}[h]
    \centering
    \includegraphics[width=0.95\linewidth]{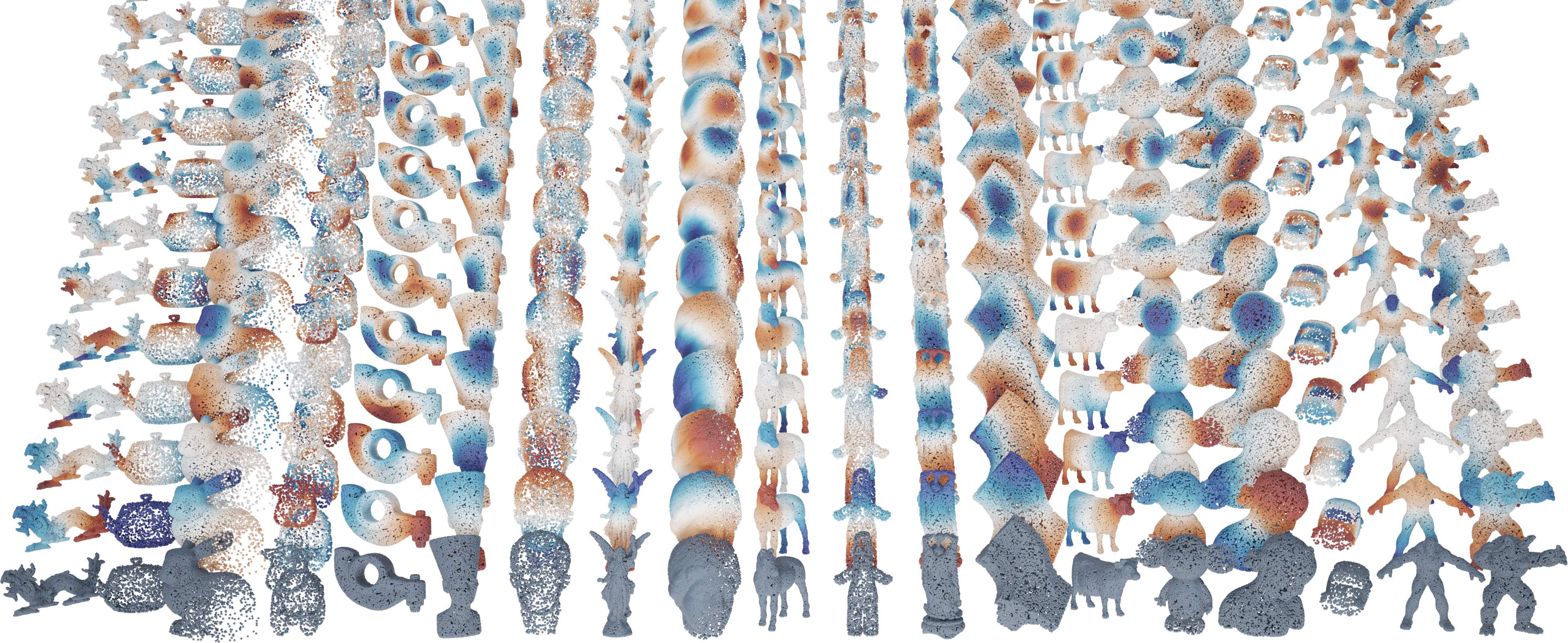}
    \caption{
    \textbf{Visualization of NEO predictions on our OOD test dataset.} The collection encompasses a wide spectrum of semantic categories, ranging from organic creatures and classic graphics models to man-made CAD parts.
    It also includes diverse topologies and structural variations, from genus-0 solids to high-genus shapes with holes and thin structures, stressing robustness beyond the training distribution.
    }
    \Description{Gallery of out-of-distribution test shapes and NEO predictions. The figure shows many models from diverse semantic categories, including animals, classic graphics models, and CAD-like objects, together with representative predicted low-frequency eigenfunctions. The collection spans different topologies, including simple solids, shapes with holes, and thin structures.}
    \label{fig:full-test-ood-gallery}
\end{figure*}

\subsection{Experimental Setup}

\paragraph{Data and discretization.}
To obtain a large and diverse training corpus, we adopt the {ShapeNetCore}~\cite{shapenet2015} dataset {($\sim$51k models, 9:1 train/test)} with a \emph{mesh-level} split to prevent data leakage from different samplings of the same geometry during training.
Unless otherwise noted, we target the first $k=96$ modes with redundancy $m=192$, and use \texttt{robust\_laplacian}~\cite{sharp2020laplacian} to construct the point cloud Laplacian and mass matrices.
To minimize data generation costs, training is performed exclusively on low-resolution point clouds ($N_{\text{train}}\!=\!2048$), where ground-truth eigenpairs are generated using ARPACK.
We further introduce {the Thingi10k dataset ($\sim$10k models) to assess cross-dataset generalization}, alongside a smaller OOD dataset of standard graphics models {(shown in \cref{fig:full-test-ood-gallery})}.
Due to the high cost of generating exact spectra at high resolutions, evaluations involving $N \gg N_{\text{train}}$ (runtime scaling and robustness) are exclusively performed on this OOD dataset.

\paragraph{Baselines, hardware, and numerical precision.}
{As a classical baseline}, we adopt ARPACK's implicitly restarted Lanczos algorithm with shift-invert~\cite{lehoucq1998arpack}.
ARPACK is evaluated in double precision on CPU (Ryzen 9 9950X, 12 threads), while NEO is evaluated on one RTX 3090 GPU with FP32/FP16 precision backbone inference and FP32 Rayleigh--Ritz.
{GPU-based iterative solvers were evaluated but performed worse than ARPACK in runtime in our setting, motivating our choice of ARPACK as the primary baseline.}
All measurements exclude $(L, M)$ construction unless explicitly stated.

\subsection{Accuracy}
\label{subsec:perf_solver}

\paragraph{Metrics}
Let $U_k$ be the ground-truth eigenvectors {provided by ARPACK} and $Y$ be the predicted basis. We evaluate performance with three metrics.
(i) \textbf{Span Loss ($\mathcal{E}_{\mathrm{span}}(i) = 1 - \|Y^\top M u_i\|_2^2$)} measures the residual energy of $u_i$ outside the predicted span. We report the mean over modes $\overline{\mathcal{E}}_{\mathrm{span}}$.
(ii) \textbf{Eigenvector MSE ($\mathcal{E}_{\mathrm{vec}}(i) = \|u_i - \hat{u}_i\|_M^2$)} measures the error of recovered vectors $\hat{u}_i$.
(iii) \textbf{Eigenvalue Error ($\mathcal{E}_{\mathrm{val}}=\frac{1}{k-1}\sum_{i=1}^{k-1}\frac{|\lambda_i-\hat{\lambda}_i|}{\lambda_i}$)} is the mean relative error of eigenvalues, excluding the trivial mode.
We report $\mathcal{E}_{\mathrm{vec}}$ and $\mathcal{E}_{\mathrm{val}}$ after the Rayleigh--Ritz procedure and mode reordering; in the presence of ambiguities, $\mathcal{E}_{\mathrm{span}}$ is the primary ambiguity-free indicator.

\begin{table}[t]
    \centering
    \caption{\textbf{Accuracy, generalization, and numerical stability across precision.}
    We report means $\pm$ std below. {SNet-32/16 and T10k-32/16 denote the ShapeNet test split and Thingi10k evaluated in FP32/FP16, respectively.}}
    \label{tab:fp_stability}

\small
\begin{tabular}{l|ccc}
\toprule
Setting & Span Loss & Evec. MSE ($\mathcal{E}_{\mathrm{vec}}$) & Eigval. Err. ($\mathcal{E}_{\mathrm{val}}$) \\
\midrule
SNet-32 & $3.7\mathrm{e}\text{-}03 \pm 1.7\mathrm{e}\text{-}03$ & $3.0\mathrm{e}\text{-}02 \pm 1.7\mathrm{e}\text{-}02$ & $3.5\mathrm{e}\text{-}02 \pm 1.4\mathrm{e}\text{-}02$ \\
SNet-16 & $3.8\mathrm{e}\text{-}03 \pm 1.7\mathrm{e}\text{-}03$ & $3.0\mathrm{e}\text{-}02 \pm 1.7\mathrm{e}\text{-}02$ & $3.6\mathrm{e}\text{-}02 \pm 1.4\mathrm{e}\text{-}02$ \\
{T10k-32} & $4.5\mathrm{e}\text{-}03 \pm 2.6\mathrm{e}\text{-}03$ & $2.3\mathrm{e}\text{-}02 \pm 1.7\mathrm{e}\text{-}02$ & $3.7\mathrm{e}\text{-}02 \pm 1.8\mathrm{e}\text{-}02$ \\
{T10k-16} & $4.7\mathrm{e}\text{-}03 \pm 2.6\mathrm{e}\text{-}03$ & $2.3\mathrm{e}\text{-}02 \pm 1.7\mathrm{e}\text{-}02$ & $3.7\mathrm{e}\text{-}02 \pm 1.8\mathrm{e}\text{-}02$ \\
\bottomrule
\end{tabular}
\end{table}

\begin{figure}
    \centering
    \includegraphics[width=\linewidth]{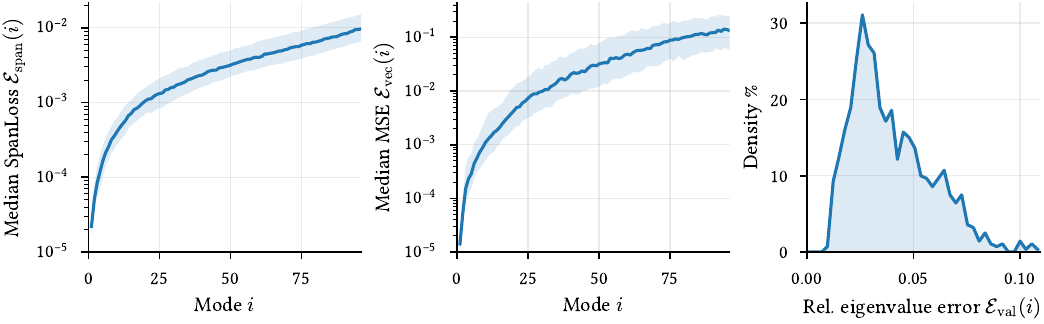}
    \caption{
    \textbf{Accuracy Distribution.}
    \textbf{Left/Middle:} Per-mode Span Loss $\mathcal{E}_{\mathrm{span}}(i)$ and Eigenvector MSE $\mathcal{E}_{\mathrm{vec}}(i)$. Solid lines denote median; shaded bands denote IQR (25--75\%).
    \textbf{Right:} Density of mean relative eigenvalue error $\overline{\mathcal{E}}_{\mathrm{val}}$.}
    \Description{Three-panel accuracy plot. The left panel shows per-mode span loss across the target spectrum, the middle panel shows per-mode eigenvector mean squared error, and the right panel shows the distribution of average relative eigenvalue error. Solid curves indicate medians and shaded regions indicate interquartile ranges.}
    \label{fig:acc_span_mse_val}
\end{figure}

\paragraph{Subspace quality and eigenpair recovery.}
We evaluate the accuracy of NEO across both the ShapeNet test set and the Thingi10k dataset.
Fig.~\ref{fig:acc_span_mse_val} (Left) shows that NEO achieves low mean span loss $\mathcal{E}_{\mathrm{span}}$, indicating the basis successfully captures the target low-frequency spectral energy.
Leveraging this subspace, Rayleigh--Ritz effectively recovers eigenpairs: the eigenvectors and eigenvalues exhibit low error (Fig.~\ref{fig:acc_span_mse_val}, Middle/Right), demonstrating consistent spectral approximation.
While errors naturally increase with frequency as the spectral gap narrows, NEO maintains robust performance across the target spectrum.
Furthermore, Table~\ref{tab:fp_stability} shows accuracy is preserved even when using half precision or {when evaluated on the Thingi10k dataset}, highlighting its numerical resilience and generalizability.

\paragraph{\texorpdfstring{Impact of subspace redundancy ($m > k$)}{Impact of subspace redundancy (m > k)}.}
We ablate the predicted subspace dimension $m$ while keeping the target rank fixed ($k=96$).
Table~\ref{tab:ablation_redundancy} shows a trade-off: a strict bottleneck ($m=k$) is too restrictive and often fails to capture spectral clusters near the cutoff.
Introducing moderate redundancy provides the necessary slack to span these modes, thereby enhancing the reconstruction quality of the eigenpairs.
Further increasing $m$ yields diminishing returns and increases the inference cost.
We therefore use $m=192$ as a practical default, balancing spectral coverage with inference latency.

\begin{table}[t]
\centering
\small
\setlength{\tabcolsep}{4.5pt}
\caption{\textbf{Effect of Subspace Redundancy ($m$) on Accuracy and Efficiency.} Target $k=96$ is fixed, and errors are reported as medians.
}
\label{tab:ablation_redundancy}
\begin{tabular}{cc|cc|c}
\toprule
\multirow{2}{*}{Output Dim $m$} & \multirow{2}{*}{Ratio $m/k$} & Span Loss & Evec. MSE & Time \\
 & & ($\overline{\mathcal{E}}_{\mathrm{span}}$) & ($\overline{\mathcal{E}}_{\mathrm{vec}}$) & (ms) \\
\midrule
96  & $1.0\times$ & 5.69e-2 & 1.21e-1 & 5.9 \\
128 & $1.3\times$ & 1.10e-2 & 6.97e-2 & 6.2 \\
160 & $1.7\times$ & 6.91e-3 & 5.50e-2 & 6.4 \\
\textbf{192 (Ours)} & \textbf{2.0$\times$} & \textbf{3.59e-3} & \textbf{3.09e-2} & \textbf{8.7} \\
256 & $2.7\times$ & 2.40e-3 & 2.15e-2 & 20.7 \\
\bottomrule
\end{tabular}
\end{table}

\subsection{Robustness}\label{subsec:robustness}
In practice, spectral pipelines often encounter distribution shifts, including changes in resolution, sampling density, and even the choice of discrete Laplacian.
We evaluate the robustness of NEO on the OOD dataset at $N=64\mathrm{k}$ ($32\times$ the training resolution), with additional stress tests up to $N=512\mathrm{k}$ as shown in Table~\ref{tab:robustness}. 

\begin{figure*}[h]
    \centering
    \includegraphics[width=0.99\linewidth]{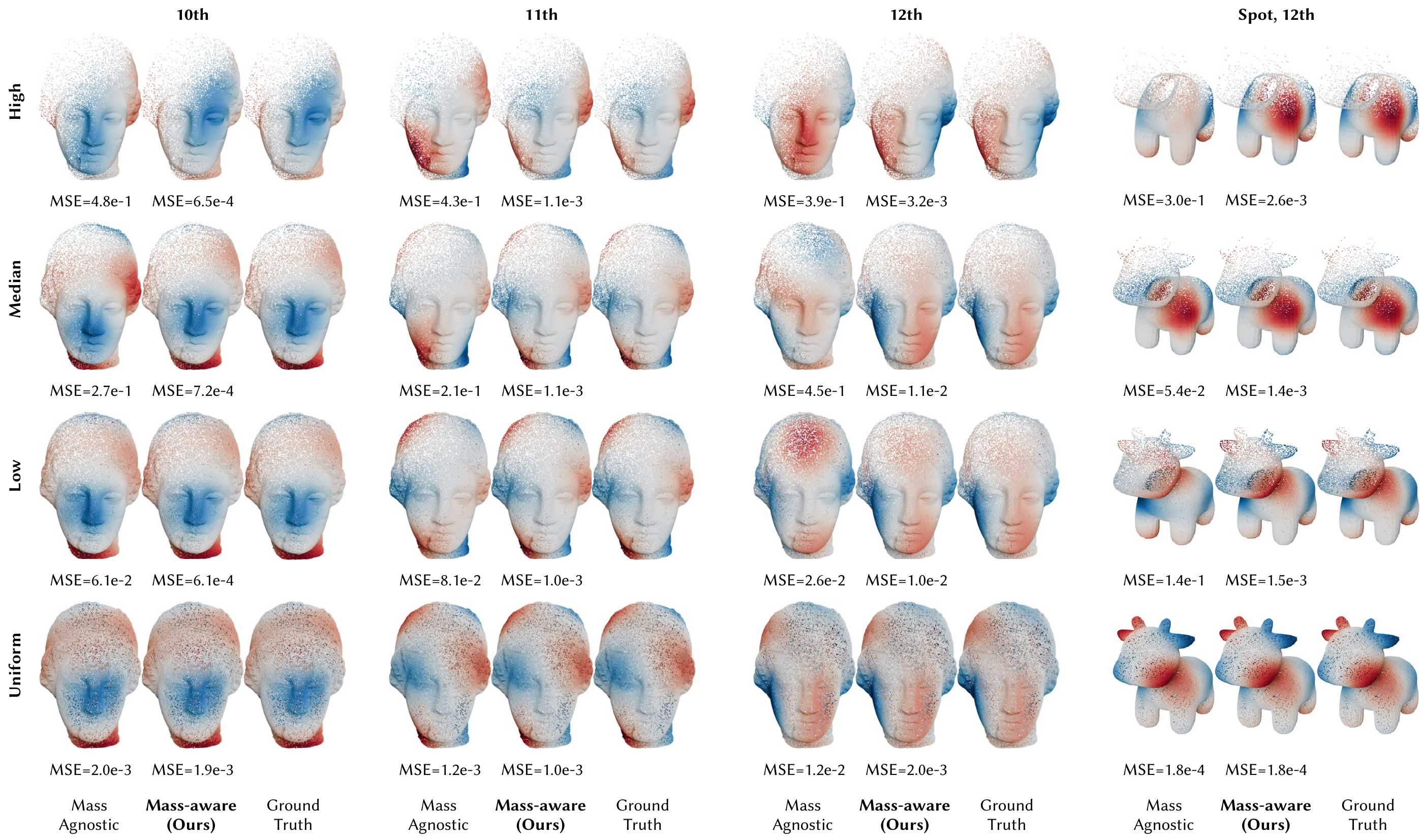}
    \caption{\textbf{Robustness to Non-Uniform Sampling.} We visualize recovered eigenvectors (10th-12th modes) under varying sampling density biases, ranging from highly non-uniform (top) to uniform (bottom). While the mass-agnostic baseline fails under biased sampling---often overfitting to high-density regions (high MSE)---our mass-aware approach consistently matches the ground truth. This confirms that injecting mass weights effectively decouples the learned features from sampling density. Notably, the mass-agnostic variant performs well on uniform data (bottom row), consistent with our analysis in \cref{subsec:backbone} that mass injection naturally reduces to standard attention when weights are constant.}
    \Description{Robustness comparison under different sampling densities. Rows show increasingly uniform point sampling, and columns compare ground-truth eigenvectors, a mass-agnostic baseline, and the mass-aware method. The mass-agnostic variant fails under strongly non-uniform sampling by concentrating on dense regions, while the mass-aware method remains visually close to ground truth across all rows.}
    \label{fig:examples-non-uniform-gallery}
\end{figure*}

\begin{figure*}
    \centering
    \includegraphics[width=0.95\linewidth]{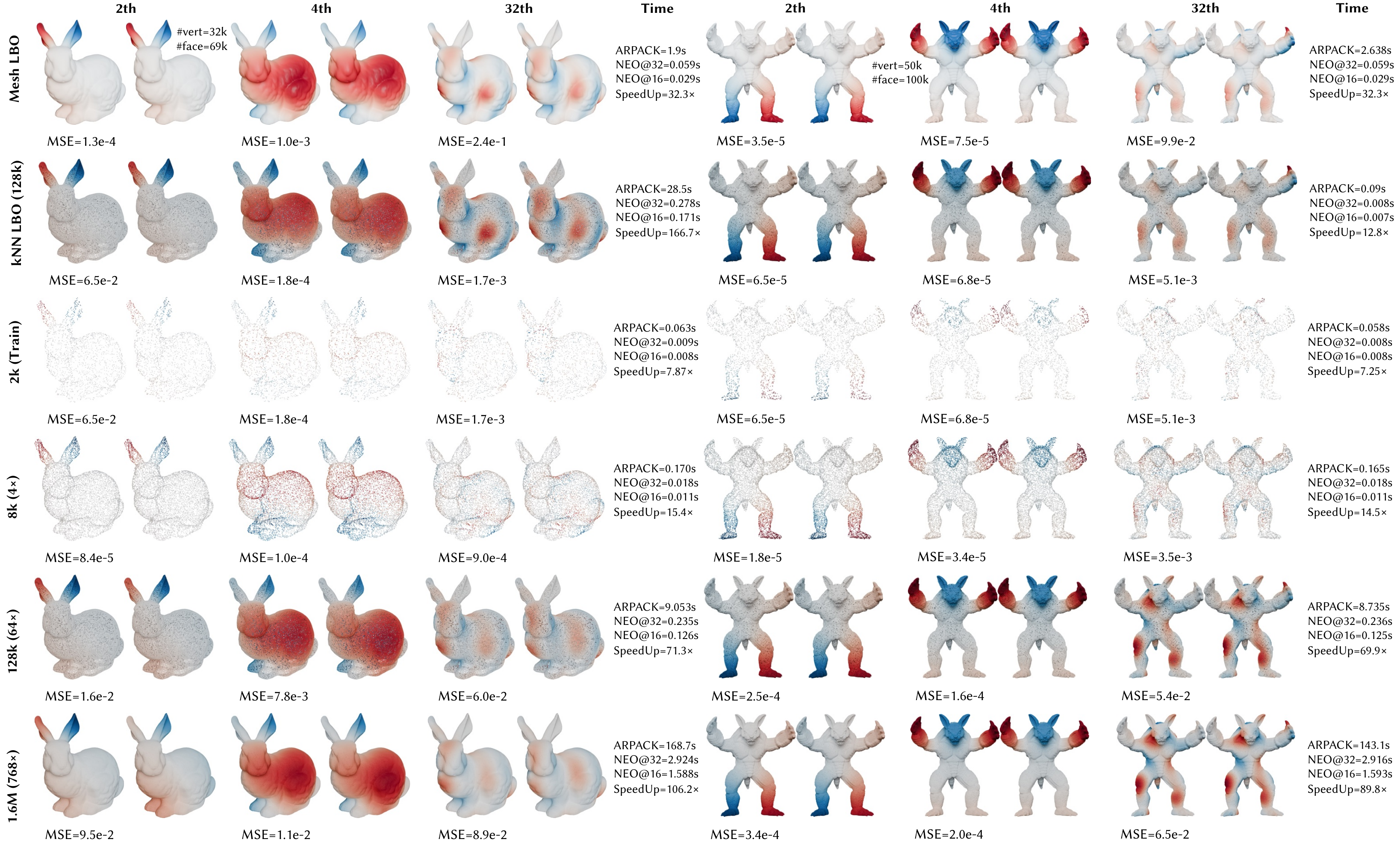}
    \caption{\textbf{Resolution scaling and discretization transfer.} We compare NEO's predicted eigenfunctions (modes 2, 4, 32) against Ground Truth across varying resolutions (2k to 1.6M points) and distinct Laplacian discretizations (Mesh vs. $k$-NN). Despite being trained only on coarse 2k point clouds, NEO demonstrates strong zero-shot generalization. While minor deviations become visible in higher frequencies (e.g., 32nd mode), the model correctly recovers the global nodal structures and qualitative patterns across all settings. The runtime columns highlight the efficiency gain: while ARPACK's cost grows super-linearly, NEO achieves over $100\times$ speedup at 1.6M points with near-linear scaling behavior.}
    \Description{Large comparison figure for zero-shot transfer across resolution and discretization. Multiple columns show ground-truth and predicted eigenfunctions for several modes at resolutions from 2 thousand to 1.6 million points, and for different Laplacian constructions such as mesh and k-nearest-neighbor graph Laplacians. Additional columns report runtimes, showing that NEO preserves global mode structure while being much faster than ARPACK at high resolution.}
    \label{fig:examples-resolution-knn-gallery}
\end{figure*}

\paragraph{Mass-awareness under non-uniform sampling.}
To isolate the effect of mass injection (Sec.~\ref{subsec:backbone}), we resample test shapes with strongly non-uniform densities and compare against a mass-agnostic variant where the mass term is removed.
As shown in Table~\ref{tab:robustness} (middle), removing mass injection makes the model highly sensitive to sampling density and leads to severe degradation.
In contrast, the mass-aware design substantially improves robustness under biased sampling, consistent with the interpretation of the down-projection attention as a measure-weighted aggregation as shown in Fig.~\ref{fig:examples-non-uniform-gallery}.

\paragraph{Zero-shot generalization to other discretizations.}

We apply NEO to distinct discrete operators (mesh cotangent and $k$-NN) without fine-tuning.
{\cref{fig:examples-resolution-knn-gallery}} and Table~\ref{tab:robustness} (bottom) show that NEO maintains strong invariance: Span Loss remains in the low $10^{-3}$ regime for both, indicating that the subspace retains most of the target energy.
This confirms that NEO learns the underlying geometry rather than overfitting to discrete artifacts, allowing the Rayleigh--Ritz step to recover eigenpairs regardless of the Laplacian choice.
{On mesh Laplacians, NEO achieves lower $\overline{\mathcal{E}}_{\mathrm{span}}$ than FastSpectrum ($1.08\times10^{-2}$) with better speedup; full comparison in Appendix A.}

\begin{table}[h]
    \centering
    \small
    \setlength{\tabcolsep}{3.5pt}
    \caption{\textbf{Robustness Analysis.} All evaluations are performed on the OOD set. 
            \textbf{Top:} Reference performance across resolutions. 
            \textbf{Middle:} Resilience to strong non-uniform sampling.
            \textbf{Bottom:} Transfer to Mesh/Graph operators.}
    \begin{tabular}{l|ccc}
    \toprule
    \multirow{2}{*}{Setting / Variant} & Span Loss & Evec. MSE & Eigval. Err.\\
    &($\mathcal{E}_{\mathrm{span}}$) &($\overline{\mathcal{E}}_{\mathrm{vec}}$) &($\overline{\mathcal{E}}_{\mathrm{val}}$) \\
    \midrule
    \multicolumn{4}{l}{\textit{Resolution Generalization (Uniform Samples)}} \\
    \quad \textbf{Reference ($N=64\mathrm{k}$)} & 3.27e-3 & 1.05e-1 & 4.59e-2 \\
    \quad Low Res. ($N=8\mathrm{k}$) & 2.58e-3 & 3.49e-2 & 1.86e-2 \\ 
    \quad High Res. ($N=512\mathrm{k}$) & 7.41e-3 & 1.71e-1 & 5.08e-2 \\
    \midrule
    \multicolumn{4}{l}{\textit{Sampling Resilience }} \\
    \quad Mass-Agnostic (Uniform) & 3.71e-3 & 1.10e-1 & 4.61e-2  \\
    \quad Mass-Agnostic (Non-Uniform) & 4.04e-1 & 3.01e-1 & 7.68e-2  \\
    \quad \textbf{Mass-Aware (Non-Uniform)} & 2.63e-3 & 8.33e-2 & 3.00e-2\\
    \midrule
    \multicolumn{4}{l}{\textit{Discretization Transfer (Uniform Samples)}} \\
    \quad {Target: Mesh LBO} & 8.65e-3 & 1.97e-1 & 3.26e-2 \\
    \quad {Target: $k$-NN LBO} & 5.67e-3 & 1.30e-1 & 8.81e-2 \\
    \bottomrule
    \end{tabular}
    \label{tab:robustness}
\end{table}

\subsection{Efficiency}

\begin{figure}[t]
    \centering
    \includegraphics[width=1\linewidth]{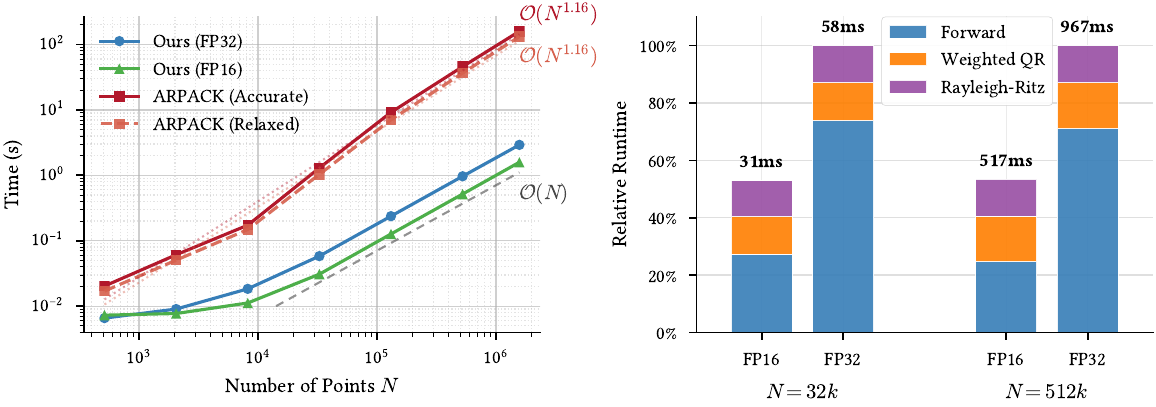}
    \caption{\textbf{Runtime analysis for $k=96$ eigenpairs.}
    \textbf{Left:} runtime scaling.
    "Accurate" and "Relaxed" indicate machine precision and matched NEO precision of ARPACK solver, respectively.
    Fit values show $t \propto N^{1.16}$ for ARPACK vs near-linear $t \propto N$ for NEO ($R^2 \ge 0.999$).
    \textbf{Right:} NEO runtime breakdown. Forward pass remains the dominant term across resolutions.}
    \Description{Two-panel runtime figure. The left panel plots wall-clock runtime against the number of input points for NEO and ARPACK under accurate and relaxed stopping criteria, showing near-linear scaling for NEO and super-linear empirical scaling for ARPACK. The right panel breaks NEO runtime into backbone forward pass, weighted QR, and Rayleigh--Ritz components at different resolutions, with the forward pass dominating total cost.}
    \label{fig:runtime}
\end{figure}

\paragraph{Runtime scaling.}
We report runtimes for recovering the first $k=96$ eigenpairs at various resolutions (Fig.~\ref{fig:runtime} left).
To evaluate ARPACK fairly, we measure it at both machine tolerance and at relaxed tolerance where the residual matches NEO's.
ARPACK exhibits super-linear empirical scaling {due to its internal sparse factorization costs}, while NEO achieves near-linear scaling.
At $N=512\mathrm{k}$, NEO recovers 96 eigenpairs in $0.52$s (FP16), yielding $88.2\times$ speedup over accurate ARPACK and a $70.2\times$ speedup over the relaxed variant.
{We evaluated several alternative solvers for comparison: LOBPCG with multigrid preconditioning failed to converge for $N > 100\mathrm{k}$, while both Spectra and GPU-accelerated SLEPc underperform ARPACK (Table~\ref{tab:alternative_solvers}), suggesting that NEO's speedup is primarily algorithmic rather than purely hardware-driven.}
{Additional solver configurations and extended timing sweeps are reported in Appendix~A.}
\begin{table}[t]
    \centering
    \small
    \caption{\textbf{Runtime Comparison of Eigensolvers.} Wall-clock time for recovering the first $k=96$ low-frequency eigenpairs at different input resolutions.}
    \label{tab:alternative_solvers}
    \begin{tabular}{l|cccc}
    \toprule
    Resolution ($N$) & SLEPc & ARPACK & Spectra & \textbf{NEO (Ours)} \\
    \midrule
    $32\mathrm{k}$  & 2.32s  & 2.01s  & 2.51s  & \textbf{31ms} \\
    $512\mathrm{k}$ & 63.7s  & 45.8s  & 60.9s  & \textbf{0.52s} \\
    \bottomrule
    \end{tabular}
\end{table}

\paragraph{Runtime breakdown.}
Fig.~\ref{fig:runtime} (right) decomposes latency into backbone forward, weighted QR, and Rayleigh--Ritz (projection + dense eigendecomposition) at $N=32\mathrm{k}$ and $512\mathrm{k}$.
The dominant cost remains the neural backbone forward pass, which scales efficiently on GPUs.
This profile confirms that increasing redundancy $m$ to improve subspace robustness incurs small latency overhead.

\section{Applications}

\subsection{Spectral Geometry Processing}
We assess whether NEO's predicted low-frequency spectrum is accurate enough for downstream spectral geometry processing tasks.

\begin{figure}[b]
    \centering
    \includegraphics[width=0.9\linewidth]{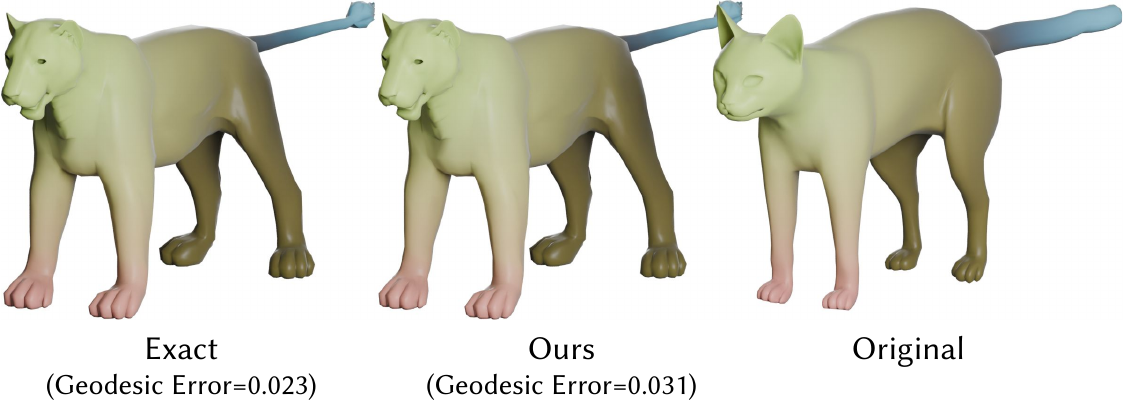}
    \caption{\textbf{Functional map correspondence.}
    Qualitative transfer results between a cat and a lion using HKS descriptors derived from exact vs. NEO eigenpairs.
    While NEO yields a slightly higher geodesic error numerically, the visual correspondence remains semantically consistent and smooth.}
    \Description{Functional-map correspondence example between a cat and a lion. The figure compares part or color transfer obtained from exact eigenpairs and from NEO-predicted eigenpairs, showing similar semantic alignment despite slightly higher numerical error for NEO.}
    \label{fig:fmap_catlion}
\end{figure}

\paragraph{Functional maps.}
We compute correspondences using the first 30 eigenpairs from NEO.
The pipeline utilizes Heat Kernel Signatures (HKS)~\cite{sun2009concise} for descriptors and ZoomOut~\cite{melzi2019zoomout} for spectral refinement.
We compare our results against maps generated using accurate spectra from a classical solver.
As shown in Figure~\ref{fig:fmap_catlion}, NEO yields visually comparable dense correspondences and preserves semantic part transfer.
{On the FAUST dataset, using NEO's predicted eigenpairs increases the mean geodesic error from 0.438 to 0.543, which remains sufficient for downstream shape matching applications.}

\begin{figure}[h]
    \centering
    \includegraphics[width=1\linewidth]{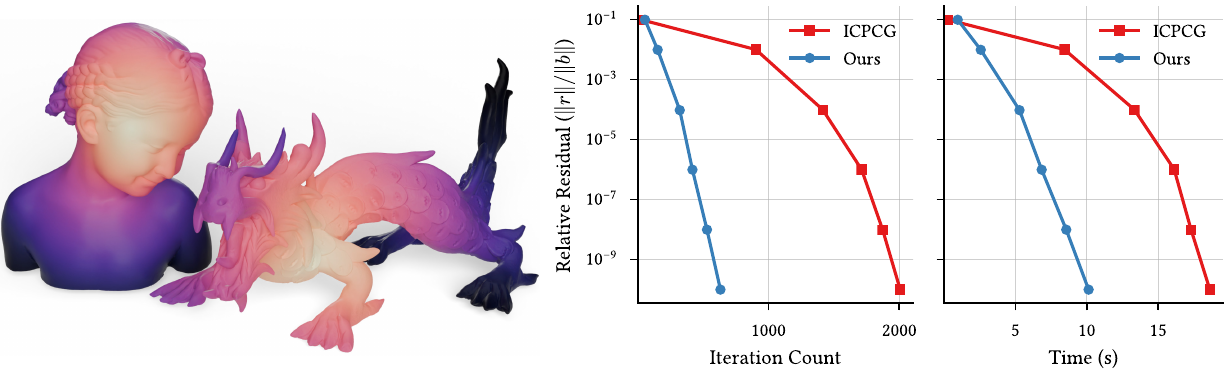}
    \caption{\textbf{Fast Poisson solve in the heat-based geodesic computation.}
    \textbf{Left:} Heat geodesic distances.
    \textbf{Middle \& Right:} Convergence of the Poisson step.
    Using NEO's predicted eigenpairs as the coarse level in an additive two-level preconditioner (Ours) reduces both the iteration count ($\sim 3\times$) and total wall-clock time compared to the standard ICPCG baseline.}
    \Description{Figure for accelerated geodesic computation. The left panel visualizes heat-based geodesic distances on a shape. The middle and right panels plot convergence behavior of the Poisson solve, comparing a standard incomplete-Cholesky preconditioned conjugate-gradient baseline with a two-level method that uses NEO eigenpairs as a coarse space. The NEO-based method reaches the target tolerance in fewer iterations and less time.}
    \label{fig:heat_convergence}
\end{figure}

\paragraph{Accelerated geodesic distances.}
We use NEO to accelerate heat-based geodesic distances~\cite{crane2017heat} by speeding up the Poisson step.
We construct an \emph{additive two-level preconditioner}: the NEO-recovered low-frequency eigenvectors form a coarse space that effectively removes slow-decaying error components, while standard incomplete Cholesky handles the fine scale.
Fig.~\ref{fig:heat_convergence} shows that this two-level approach significantly reduces both iteration count and wall-clock time compared to the ICPCG baseline, demonstrating that our predicted subspace is high-quality enough to accelerate classical numerical solvers.


\subsection{Point Embeddings without Explicit Spectra}
We investigate whether the raw predicted subspace $F$ can serve as a robust intrinsic embedding, bypassing the eigensolving stage.
We posit that low-frequency invariant subspaces naturally encode global intrinsic information, serving as a powerful geometric prior to facilitate downstream geometric learning.

\begin{figure*}[t]
    \centering
    \begin{subfigure}[t]{0.5\linewidth}
        \centering
        \includegraphics[width=\linewidth]{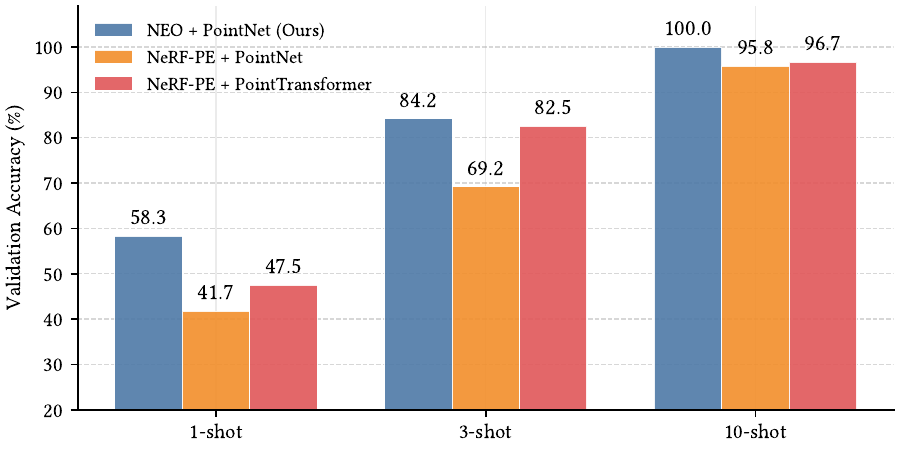}
        \caption{\textbf{Few-shot classification (SHREC-11).} Validation accuracy for 1-shot, 3-shot, and 10-shot settings (higher is better).}
        \Description{Few-shot classification plot on SHREC-11 showing validation accuracy for 1-shot, 3-shot, and 10-shot settings. Curves or bars compare a lightweight PointNet head using frozen NEO features against baselines such as NeRF-style positional encoding and Point Transformer, with NEO achieving the highest accuracy, including perfect accuracy in the 10-shot case.}
        \label{fig:emb_cls}
    \end{subfigure}\hspace{0.07\linewidth}
    \begin{subfigure}[t]{0.37\linewidth}
        \centering
        \includegraphics[width=\linewidth]{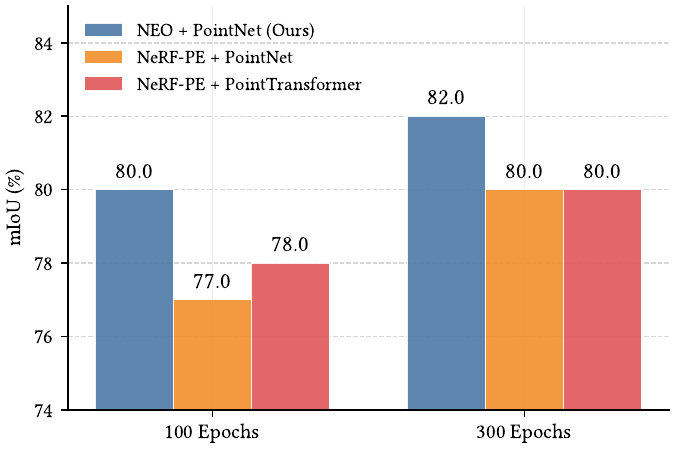}
        \caption{\textbf{Segmentation (Human).} Mean Intersection over Union (mIoU) at 100 and 300 epochs (higher is better).}
        \Description{Segmentation performance plot on the human dataset showing mean intersection over union at 100 and 300 training epochs. The comparison indicates faster convergence and better accuracy when using frozen NEO features than when learning directly from positional encoding under the same budget.}
        \label{fig:emb_seg}
    \end{subfigure}
    \caption{\textbf{NEO as an intrinsic point embedding.}
    We use the raw predicted subspace $F$ directly as point features without explicit eigensolving, comparing it against NeRF-style positional encoding (NeRF-PE) baselines.
    \textbf{(a)} For classification, frozen NEO features enable a lightweight PointNet to achieve perfect accuracy, surpassing heavier end-to-end baselines (e.g., PointTransformer).
    \textbf{(b)} For dense segmentation, NEO's global spectral features significantly accelerate convergence compared to learning from spatial coordinates under the same training budget.
    }
    \Description{Two-panel figure showing NEO as an intrinsic point embedding. Panel (a) reports few-shot classification accuracy on SHREC-11 for several training-shot regimes and shows that frozen NEO features outperform baseline point encodings and heavier end-to-end models. Panel (b) reports segmentation mean intersection over union at two training checkpoints and shows that NEO features improve convergence and final accuracy.}
    \label{fig:embeddings}
\end{figure*}

\paragraph{Few-shot classification.}
We evaluate on SHREC-11~\cite{lian2011shrec} (official split, 30 classes).
Here, NEO functions as a \emph{frozen} feature extractor: we fix the backbone and train only a lightweight PointNet~\cite{qi2017pointnet++} head on top of $F$.
We compare against training from scratch using NeRF-style positional encodings (NeRF-PE) with the same head, and a much heavier Point Transformer~\cite{zhao2021point} baseline.
Fig.~\ref{fig:embeddings} (a) reveals a striking advantage: frozen NEO features enable a simple PointNet to reach \textbf{100\% accuracy} (10-shot), significantly outperforming the Point Transformer.
This suggests that NEO provides an effective intrinsic representation, reducing the need for complex architectural designs.

\begin{figure}[h]
    \centering
    \includegraphics[width=0.99\linewidth]{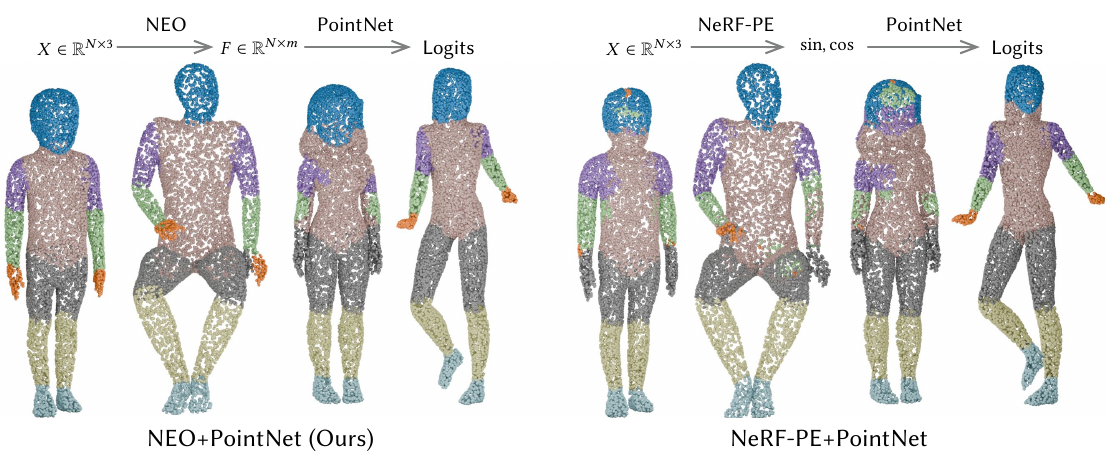}
    \caption{\textbf{Segmentation.}
Visual comparison of segmentation results using our predicted basis ($F$) versus sinusoidal positional encoding (NeRF-PE).
    }
    \Description{Qualitative segmentation comparison on human shapes. The figure contrasts predictions obtained using NEO-derived point features with predictions using sinusoidal positional encoding, showing that NEO better separates semantically distinct but spatially adjacent body parts.}
    \label{fig:segmentation}
\end{figure}

\paragraph{Segmentation.}
We further assess dense prediction on the human body segmentation task~\cite{maron2017convolutional} under a fixed training budget.
Using the same setup, we train a segmentation head on top of frozen $F$ versus learning from coordinates.
As shown in Fig.~\ref{fig:embeddings} (b), the spectral prior provided by $F$ yields faster convergence and higher mIoU under the limited training budget.
While end-to-end models eventually narrow the gap, NEO provides explicit {global information} that is otherwise hard to learn from local neighborhoods (see Fig.~\ref{fig:segmentation}), significantly reducing optimization difficulty.

